\documentclass{article}

\usepackage{microtype}
\usepackage{graphicx}
\usepackage{subfigure}
\usepackage{multirow}
\usepackage{booktabs} %

\usepackage{hyperref}

\usepackage[accepted]{icml2025}

\usepackage{amsmath}
\usepackage{amssymb}
\usepackage{mathtools}
\usepackage{amsthm}

\usepackage[capitalize,noabbrev]{cleveref}

\theoremstyle{plain}
\newtheorem{theorem}{Theorem}[section]

\newtheorem{corollary}[theorem]{Corollary}
\theoremstyle{definition}

\theoremstyle{remark}

\usepackage[textsize=tiny]{todonotes}

\usepackage{caption, subcaption}

\newcommand{\papername}{Make Optimization Once and for All with Fine-grained Guidance}
\newcommand{\methodname}{Diff-L2O }
\newcommand{\methodnameE}{Diff-L2O}

\icmltitlerunning{\papername
}
\usepackage{colortbl}
\usepackage{xcolor}

\begin{document}

\newcommand{\SamJ}[1]{\textcolor{red}{#1}}

\newlength\savewidth\newcommand\shline{\noalign{\global\savewidth\arrayrulewidth
  \global\arrayrulewidth 1pt}\hline\noalign{\global\arrayrulewidth\savewidth}}

\definecolor{defaultcolor}{HTML}{E8E2F7}
\newcommand{\default}[1]{\cellcolor{defaultcolor}{#1}}

\definecolor{baselinecolor}{gray}{.9}
\newcommand{\baseline}[1]{\cellcolor{baselinecolor}{#1}}

\newcommand{\tablestyle}[2]{\setlength{\tabcolsep}{#1}\renewcommand{\arraystretch}{#2}\centering\footnotesize}
\renewcommand{\paragraph}[1]{\vspace{1.25mm}\noindent\textbf{#1}}

\twocolumn[
\icmltitle{\papername}

\icmlsetsymbol{equal}{*}

\begin{icmlauthorlist}
\icmlauthor{Mingjia Shi}{equal,nus}
\icmlauthor{Ruihan Lin}{equal,hkust}
\icmlauthor{Xuxi Chen}{uta}
\icmlauthor{Yuhao Zhou}{nus}
\icmlauthor{Zezhen Ding}{hkust}
\icmlauthor{Pingzhi Li}{unc}
\icmlauthor{Tong Wang}{unc}
\\
\icmlauthor{Kai Wang}{nus}
\icmlauthor{Zhangyang Wang}{uta}
\icmlauthor{Jiheng Zhang}{hkust}
\icmlauthor{Tianlong Chen}{unc}
\end{icmlauthorlist}

\icmlaffiliation{nus}{National University of Singapore}
\icmlaffiliation{hkust}{The Hong Kong University of Science and Technology}
\icmlaffiliation{uta}{University of Texas at Austin}
\icmlaffiliation{unc}{University of North Carolina at
Chapel Hill}

\icmlcorrespondingauthor{Tianlong Chen}{tianlong@cs.unc.edu}

\icmlkeywords{Machine Learning, ICML}

\vskip 0.3in
]

\printAffiliationsAndNotice{\icmlEqualContribution} %

\begin{abstract}
Learning to Optimize (L2O) enhances optimization efficiency with integrated neural networks. L2O paradigms achieve great outcomes, \textit{e.g.}, refitting optimizer, generating unseen solutions iteratively or directly. However, conventional L2O methods require intricate design and rely on specific optimization processes, limiting scalability and generalization. Our analyses explore general framework for learning optimization, called \textit{Diff-L2O}, focusing on augmenting sampled solutions from a wider view rather than local updates in real optimization process only. Meanwhile, we give the related generalization bound, showing that the sample diversity of Diff-L2O brings better performance. This bound can be simply applied to other fields, discussing diversity, mean-variance, and different tasks.
Diff-L2O's strong
compatibility is empirically verified with only minute-level
training, comparing with
other hour-levels. 
\end{abstract}

\section{Introduction}

Learning to optimize (L2O)~\citep{chen2017learning,chen2022scalable,metz2022velo,li2016learning} aims to improve the efficiency of optimization algorithms by refitting optimization algorithms with (machine) learning. Learning optimization algorithms involved in iteration, it has significant advantages in accelerating optimization algorithms~\citep{chen2022learning,xie2024ode,zheng2022symbolic,cao2019learning}.

Among the objectives of L2O speedup optimization algorithms are usually composed of the following paradigms with great performance. 1) Learning the settings of the optimizer so as to ~\citep{xie2024ode} find a set of settings that make the optimizer search the solution space faster and more stable; 2) using a generator to guide the model iteration, e.g., iterating the model step by step with the inference of a sequence model ~\citep{chen2017learning}; 3) modeling the parameter space directly and generating the parameters of the model in a better way~\citep{gartner2023optimus}.

However, L2O methods require delicate design and tuning, depending on real optimization processes. These paradigms 1) do not directly model the optimization process in general but each point on trajectories or 2) rely on the real optimization process of specific types of optimizers. These facts limit L2O scaling up~\citep{metz2022velo}, and loss the advantage of the generalization capabilities brought by machine learning. Ours below helps solving potentially unknown optimization problems w/o sophisticated designs.

\begin{figure}
    \centering 
    \includegraphics[width=0.98\linewidth]{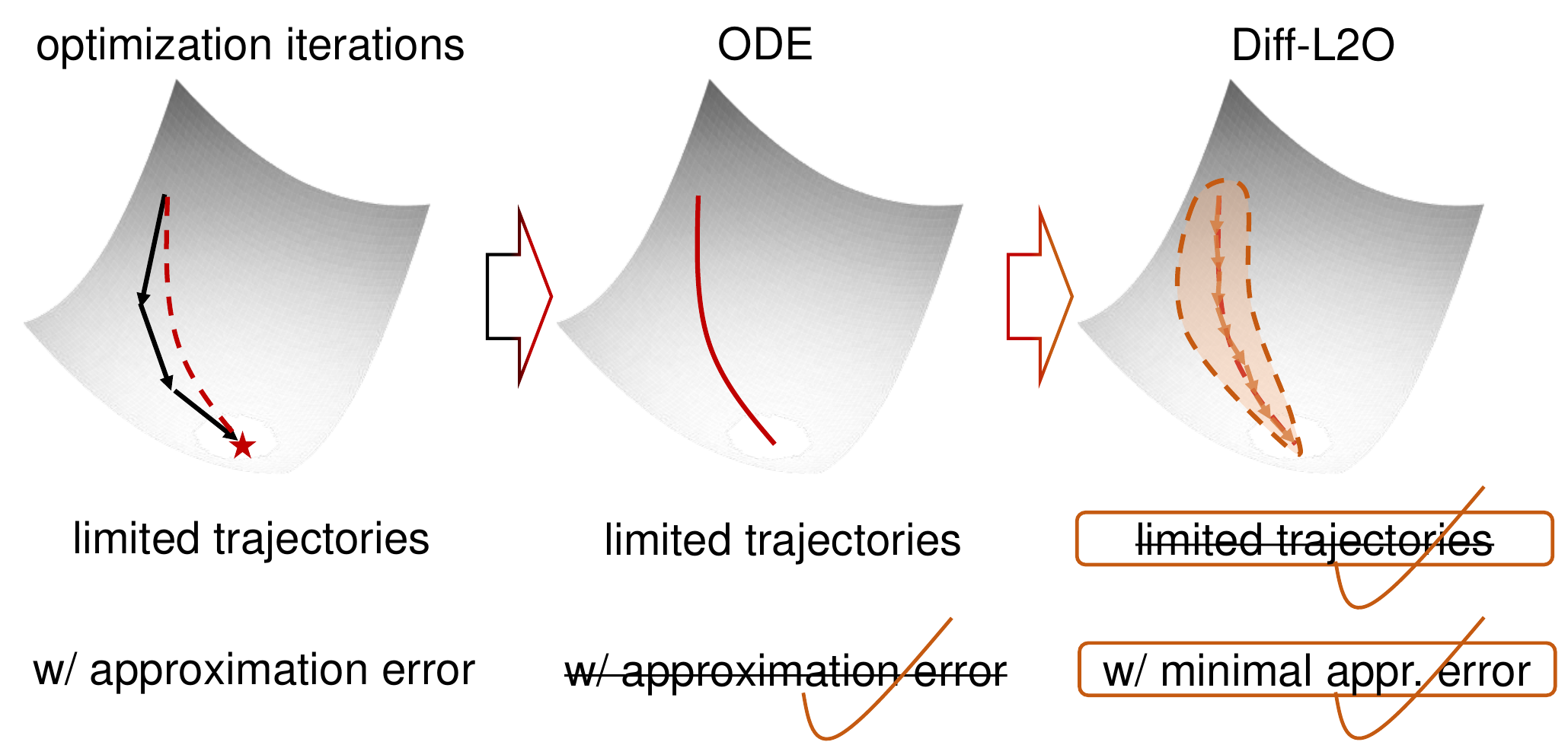}
    \vspace{-10pt}   
    \caption{Diff-L2O's intuitions: wider views and better sampling diversity with artificial sampling on solution spaces.}
    \vspace{-20pt}
    \label{fig:fig_intro}
\end{figure}

Corresponding to the two aforementioned points respectively, discussion is about 1) the feasibility of unified modeling~\citep{attouch2019first} for the vast majority of optimization algorithms, and the corresponding optimizers, by means of unified modeling~\citep{xie2024ode}; 2) propose a optimization with wider views, \textit{i.e.}, find a range to the solution, rather than finding a locally best update direction. We explore the generalization performance under this unified modeling and give the generalization bound. We brief the main analyses that \textit{augmentation with diffusion improves generalization} of the modeled solutions.

Empirically, the proposed Diff-L2O demonstrates adaptability to quickly obtain initial points and further speed-up optimization for classic optimizers. Only second-level training time cost are needed for Diff-L2O, comparing with other hour-level methods. It also works on deep neural networks.\footnote{Results on DNN are in the Appendix.}

The contributions of this work are as follows:
\begin{itemize}
    \item We propose a fast method for solving optimization problems using diffusion models while combining artificial and real data with guidance information.
    \item We analyze the key factors that can be used to model the solution space with generative models, as well as general formulation, and related generalization bound.
    \item Experiments using diffusion models to model the solution space, thus accelerating optimization, have yielded impressive results with the proposed Diff-L2O.
\end{itemize}
\section{Methodology}
\subsection{Preliminary}
\paragraph{Optimization's general trajectory formulation.}
The dynamics of the vast majority of optimization methods, known as the Inertial System of Hessian-driven Damping~\citep{attouch2019first} (ISHD), can be represented as following:

\begin{equation}
\label{equ:mtd_ISHD}
\ddot{x}+\frac{\alpha}{t}\dot{x}+\beta\nabla^{2}f(x)\dot{x}+\gamma\nabla f(x)=0\text{,}
\end{equation}

where $\nabla$ and $\nabla^{2}$ are the gradient and Hessian operations respectively, $\dot{x}$ and $\ddot{x}$ are the first and second ordered derivatives of $x$ on time $t$, and, $\alpha$, $\beta$ and $\gamma$ are hyperparameters on $t$ (which abbreviates $\alpha_{t}, \beta_{t}, \gamma_{t}$) that determine the trajectories of the optimization algorithms.

In L2O cases, we want to learn the solution space of the problem $\min_{x} f(x)$. The model is actually approximating the ODE (\textit{i.e.}, the $\alpha$, $\beta$ and $\gamma$).

\paragraph{Discretization.}
Euler discretization is an efficient and commonly used discretization method. It is primarily affected by non-linear sampling scenarios. In such cases, the rugged and unknown real optimization surface limits the possibility of further acceleration~\cite{xie2024ode,schuetz2022combinatorial} and can easily lead unstable results.

\paragraph{Stochastic optimization's dynamics.} The dynamics in Equ.~\ref{equ:mtd_ISHD} is the general ODE of the most gradient-based optimization trajectories. However, more practical dynamics are stochastic ones, which can be represented by stochastic differential equations. The Ito formula of Wiener process the SDE can be presented as follows.
$$\mathbf{d}\tilde{x}=u\mathbf{d}t+v\mathbf{d}w\text{,}$$
where $w$ is the Brownian motion, $u$ and $v$ are the functions on $t$ determining the types, which abbreviates $u_{t}$ and $v_{t}$.

\paragraph{Diffusion process.}
The aforementioned classic formulation of a diffusion process is not enough since due to direct expression of different common stochastic processes. So we have the following more specific ones.
In a more general case, we reformulate it into the following one.
\begin{equation}
\mathbf{d} \tilde{x} = \tilde{x}\dot{s} /s \mathbf{d} t + s \sqrt{\dot{\sigma}\sigma}\mathbf{d} w,~ \tilde{x}=s\tilde{x}_{0}+s\sigma\epsilon, \epsilon \sim \mathcal{N}(0,\mathbf{I})\text{,}
\label{equ:mtd_pre_diffusion_process}
\end{equation}
where $\sigma$ and $s$ abbreviates $\sigma_{t}$ and $s_{t}$, $\tilde{x}_{t}$ is the stochastic process with given $\tilde{x}_{0}$ as initial point.

\subsection{Discussion: modeling solutions is feasible}

We intuitively give a discussion about the main cases and our motivation. See Sec.~\ref{sec:mtd_method_analyses} for more details.

\paragraph{Takeaways.} Our discussion is summarized below.
\\
1) Optimization process's meta features do provide information for solution space modeling;
\\
2) The data from real optimization process is helpful, but it is still not enough.

\paragraph{Case: overparameterization.}
We know that optimization algorithms have their own implicit biases (or, regularization)~\citep{gunasekar2018characterizing}, when the case goes with overparameterization, \textit{e.g.}, small norms, sparse solutions, flat (stable) solutions, small gradients, maximum margin.

The implicit biases~\citep{dauber2020can,soudry2018implicit,gunasekar2018implicit} depend on the problem formulation and the optimization algorithm.
which means that the \textit{optimization formulation and algorithm is informative} to the expected results. Linear regression, for example, tends to a min-norm solution with gradient descent optimizer.

\paragraph{Case: underparameterization.}
The implicit biases within under-parameter classical problems~\citep{bowman2022implicit} can be reduced into non-under-parameter cases. For example, linear regression can be full-ranked on subspaces while maintaining the similar solution spaces and keeping the form of implicit bias.

\paragraph{Case: low performance.} Moreover, low performance in the under-parameterized case would not directly related to the feasibility of solution spaces being modeled. It would make the surface more mundane and some SDEs more chaotic. That is, while performance is low, and yet the parameter space is easy to approximate, because the prediction only need to be noise, given targeted chaotic SDE.

Thus, the optimizer, the problem itself (\textit{i.e.}, optimizee), and other meta-features are informative about the solution generation. The all will help modeling the probabilistic (generative) model of the solution spaces in details.

\paragraph{Closest doesn't mean best.} Different implicit biases imply different probability distributions of solutions. Unexplored implicit biases could bring better solutions within the solution space. The closest approximations to the trained solutions or the converged SDEs are thus not the best. Decoupling dependency on real optimization trajectories is greater potential for generalization.

\paragraph{The closest is yet informative.} Well-fit-SDE models can still tell us a lot. For example, in the case where \textit{mode connectivity}~\citep{garipov2018loss} is considered, the terminal phases of the optimization SDEs do not exactly converge, but rather swim around within a connected region toward the similar-performance region that meets the implicit bias.

Thus, modeling the solution space with great diversity and guidance with real information are both required.

\subsection{Diff-L2O: How to model solutions}
According to the discussion, our approach focuses on 1) using intermediate quantities from the optimization process as guidance, while, 2) using both real and artificial SDE to ensure roughly validity while exploring unseen solutions.

\paragraph{Artificial trajectories: diffusion process.}
Global optimals are costly. Random noise is introduced to explore more solutions. It, meanwhile, need to be close to given SDE, in order to make full use of the information available. Diffusion processes are started from some suboptimal solutions. It maintains the continuity and smoothness of the random trajectory. Meanwhile, the orientation is towards potentially possible solutions.

The diffusion process is simulated according to the current big-hit diffusion models. These models simulate and fill the designed parameter space very well, mainly involved in the following $s$-$\sigma$ forms~\citep{karras2022elucidating}. The diffusion processes' general forms are shown in Equ.~\ref{equ:mtd_pre_diffusion_process} and specialized in Tab.~\ref{tab:diffusion_process}, including DDPM (VP-SDE)~\citep{ho2020denoising}, VE-SDE~\citep{song2021score} and EDM\citep{karras2022elucidating}.

\begin{table}[t]
\tablestyle{9pt}{1.3}
    \centering
    \caption{The ingredients of mainstream SDE designs.}
    \begin{tabular}{cccc}

        SDEs
        &
        VP
        &
        VE
        &
        EDM
        \\
        \shline
        $s$
        &
        $\exp\{-\frac{1}{4}\Delta_{\beta}t^{2}-\frac{1}{2}\beta_{0}t\}$
        &
        $1$
        &
        $1$
        
        \\
        $\sigma^{2}$
        &
        $\exp\{\frac{1}{2}\Delta_{\beta}t^{2}+\beta_{0}t\}-1$
        &
        $t$
        &
        $t^2$

        \\
        $\dot{s}$
        &
        $-\frac{\sigma\dot{\sigma}}{(1+\sigma^{2})^{3/2}}$
        &
        $0$
        &
        $0$
        \\
        $\dot{\sigma}$
        &
        $\frac{(1+\sigma^{2})(\Delta_{\beta}t+\beta_{0})}{2\sigma}$
        &
        $1$
        &
        $2t$
        \\
        \multicolumn{4}{r}{\text{
        $\vartriangleright~$$\beta_{0}$ and $\Delta_{\beta}$ are pre-defined parameters.}}
    \end{tabular}
    \vspace{-0pt}
    \label{tab:diffusion_process}
\end{table}

\paragraph{Discretization and sampling.}
For sampling on the SDE of the diffusion process, we use the simple and efficient Euler sampling. The SDE is isotropic diffusion using DDPM (VP-SDE)~\citep{ho2020denoising, song2021score}.

Besides sampling efficiency, it lies in the existence of an implicit L2-norm term in our subsequent approach. The isotropic parameter space design and variance preserving DDPM are fitted just right. The samling algorithm are shown in Algorithm~\ref{alg:mtd_diff_fwd} and Algorithm~\ref{alg:mtd_diff_bkw}.

\begin{algorithm}[t]
\caption{Forward Scheduling}
\label{alg:mtd_diff_fwd}
\textbf{Inputs:} The starting point of the forward trajectory $\tilde{x}_0$, and a coefficient list $[\Bar{\alpha}_0, \dots, \Bar{\alpha}_\mathtt{T}]$ 
\begin{algorithmic}
\FOR{$t = 1, 2, \dots, \mathtt{T}$}
    \STATE $\tilde{x}_{t} \gets \mathcal{N}(\sqrt{\Bar{\alpha}_t} \tilde{x}_0, (1-\Bar{\alpha}_t)\mathbf{I})$
\ENDFOR
\end{algorithmic}
\textbf{Output:} $[\boldsymbol{x}_0,\boldsymbol{x}_1,\dots,\boldsymbol{x}_\mathtt{T}]$
\end{algorithm}

\begin{algorithm}[t]
\caption{Backward Sampling}
\label{alg:mtd_diff_bkw}
\textbf{Inputs:} A standard Gaussian noise $\boldsymbol{\hat{x}}_\mathtt{T}\sim\mathcal{N}(0, \mathbf{I})$, and a guidance vector $\boldsymbol{g}$. 
\begin{algorithmic}
\FOR{$t = \mathtt{T}, \mathtt{T}-1, \dots, 1$}
    \STATE $\boldsymbol{t} \gets \texttt{TE}(t)$
    \STATE $\boldsymbol{\hat{x}}_{t-1} \gets \texttt{opt}(\texttt{concat}(\boldsymbol{\hat{x}}_t, \boldsymbol{g}, \boldsymbol{t}))$
\ENDFOR
\end{algorithmic}
\textbf{Output:} $\boldsymbol{\hat{x}}_0$
\end{algorithm}

\paragraph{Training: Diff-L2O.}
Since our approach is Euler sampling on VP-SDE, we use $\epsilon$-parameterization to train our diffusion model, according to DDPM. However, DDPM does not consider how the solution behaves in the optimization process, only whether it is aligned well with white noise.

Relatively, our approach takes that into account. Our approach uses the aforementioned guidance (\textit{e.g.}, quantities in the processes, optimization meta-features). These help the parameter space modeled to be embedded with meta-information about optimization. This brings greater generalizability. Meanwhile, we add the loss of the current solutions on the optimization objective as a metric that is integrated uniformly into the probabilistic modeling of the generated model. The algorithm is shown in Algorihtm~\ref{alg:mtd_diff_diffl2o_training}

\begin{algorithm}[t]
\caption{Diff-L2O Training}
\label{alg:mtd_diff_diffl2o_training}
\textbf{Inputs:} Initial point ${\hat{x}}_\mathtt{T}\sim\mathcal{N}(0, \mathbf{I})$, guidance vector ${g}$, the optimizee's parameter ${\theta}$, the forward trajectory $\{\tilde{x}_0,\tilde{x}_1,\dots,\tilde{x}_\mathtt{T}\}$, loss coefficient $\alpha$  
\begin{algorithmic}
    \FOR{$t = \mathtt{T}, \mathtt{T}-1, \dots, 1$}
        \STATE $\boldsymbol{t} \gets \texttt{TE}(t)$
        \STATE ${\hat{x}}_{t-1} \gets \texttt{opt}(\texttt{concat}({\hat{x}}_t, {g}, \boldsymbol{t}))$
        \STATE $\mathcal{L}_1 \gets  f({\theta}, {\hat{x}}_{t-1})$
        \STATE $\mathcal{L}_2 \gets  \text{MSE}(\tilde{x}_{t-1}, {\hat{x}}_{t-1})$
        \STATE $\mathcal{L} \gets \alpha \mathcal{L}_1 + (1-\alpha) \mathcal{L}_2$ 
        \STATE Update \texttt{opt} by minimizing $\mathcal{L}$
    \ENDFOR
\end{algorithmic}
\end{algorithm}

\paragraph{Generalization analyses.}
\label{sec:mtd_method_analyses}
Diff-L2O augments the diversity of the samples and hence works better. The relevant theorem  on our setting is from the perspective of PAC-Bayesian.

The generalization gap is defined as following:

\begin{align*} 
\Delta(\hat{x})&:=\Delta(\hat{f}_{S},\hat{f}_{D})\text{,}
\\
&\text{~where~}\hat{f}_{\cdot}\text{~abbr.~}f(\hat{x};\cdot):=\mathbf{E}_{d\sim\cdot}{f}(\hat{x};d)\}\text{.}
\end{align*}
$\hat{f}_{\cdot}$ and $f_{\cdot}$ are the problems' expectation values of $\hat{x}$ and $x$ on probability from approximated model $q$ or the real solution space distribution
(\textit{w.r.t.}, $\min$ for simplification), $D$ and $S$ are the population (test) and samples  (train), \textit{i.e.}, ground truth and sampled solutions in L2O. $\Delta$ abbr. distance $\Delta(\hat{x})$.

This differs the previous PAC-Bayesian bounds in the artificial samples' distribution and $\hat{x}_{t}\sim q_{t}(g)$ obtained from a stochastic process of guidance $g$, $\textit{e.g.}$, meta-features. The time $t$ and condition $g$ are omitted for simplicity below.

\begin{theorem}(General PAC-Bayesian on stochastic solution space.) In this general theorem, $\Delta$ requires only a non-negative general convex distance, and we do not restrict the optimization objective to the downstream tasks. With a initial prior process $p$, $\forall q$ (posterior) w/ $n$ \#samples, we have the following bound at least $1-\delta$ probability:
$$\Delta \le_{1-\delta} \frac{1}{n}\{\mathrm{KL}(q||p)+\log \frac{\mathcal{M}}{\delta}\},\forall \text{time~}t$$
where $\mathcal{M}:=\mathbf{E}_{h\sim p}\exp\{n\Delta(h)\}$ is related to the optimization task, including the distance between population and the training set.
\end{theorem}
\begin{proof}
    With given probability $1-\delta$ (w.h.p.), we have $$\Delta(\hat{f}_{S},\hat{f}_{D})\le\epsilon_{\delta}(n)\text{.}$$
    As our problem is defined as $\min$ for simplification, we focus on the upper bound here.

    From the expectation extended objective functions: $\hat{f}_{D}=\mathbf{E}_{\hat{x}\sim q}\Delta$ and $\hat{f}_{S}=\mathbf{E}_{\hat{x}\sim q}f(\hat{x}; S)$, we decouple a prior $p$ from modeled distribution $q$ with Jensen inequality,
    $$\log \mathbf{E}_{h\sim p} \exp\{n\Delta(h)\}\ge n\Delta-\mathrm{KL}(q||p)$$

    With Markov inequality, introducing probability 1-$\delta$,

    $$\Delta \le \frac{1}{n}\{\mathrm{KL}(q||p)+\log \frac{\mathcal{M}}{\delta}\}\text{,~w.h.p.,}$$
    where $\mathcal{M}:=\mathbf{E}_{h\sim p}\exp\{n\Delta(h)\}$ is independent of $q$. It should be discussed in different optimization objectives and downstream tasks. The all do not depend on time $t$ here.
\end{proof}

General generalization upper bounds are time-independent, and next we discuss specific SDE modeling processes that are time-dependent, and their relationship to tasks.

\begin{corollary}(Diff-L2O: Gaussian.) When $p\sim\mathcal{N}(\mu, \Sigma)$, $q\sim\mathcal{N}(\hat{\mu}, \hat{\Sigma})$, the KL-divergence is $$\mathrm{KL}(q||p):=\frac{1}{2}\{\log\frac{|\Sigma|}{|\hat{\Sigma}|}-k+||\hat{\mu}-\mu||^{2}_{\Sigma}+\mathrm{tr}(\Sigma^{-1}\hat{\Sigma})\}\text{.}$$
In Diff-L2O, the Gaussian is isotropic, and initial prior $p \sim \mathcal{N}(\sqrt{\bar{\alpha}_{t}} x, (1-\bar{\alpha}_{t})\mathbf{I})$, $x\sim D$. We can further format the bound as 
\begin{align*}
\Delta&\leq_{1-\delta} \frac{1}{n}\{k\log(1-\bar{\alpha}_{t})-\log|\hat\Sigma|-k+||\hat{\mu}-\mu||^{2}_{2}
\\
&+\frac{\mathrm{tr}(\hat{\Sigma})}{(1-\bar{\alpha})}+\log\frac{\mathcal{M}}{\delta}\} 
\text{,~where~} k=\mathrm{dim}~x\text{.}
\end{align*}.
\end{corollary}

\begin{corollary} (Diff-L2O: Classification tasks.)
Generalizing over the classification task, we define $\hat{f}_{D}$ and $\hat{f}_{S}$ by considering the prediction error rate of the modeling probability $q$ on the test and training sets, and use the difference between the two as the distance $\Delta$.

If the error rate is $m/n$ ($m$ misclassified samples among $n$ samples), we have the probability:
$$\mathbf{P}_{\tilde{S}\sim D}(\hat{f}_{S}={m}/{n})=\mathrm{Bio}(m;n, \hat{f}_{D}),\forall m\text{,}$$
where $\tilde{S}$ is a set of $m$ independent samples. We have:
$$\mathcal{M}=\sup_{\mathcal{P}\in[0,1]} [\sum_{m=0}^{n}\mathrm{Bio}(m;n,\mathcal{P})\exp\{n\Delta(m/n,\mathcal{P})\}]$$
\end{corollary}

Thus, we have the following bound, when Diff-L2O is applied to general classification tasks or other tasks can be reduced into classification.
\begin{align*}
&\Delta \leq_{1-\delta}\\
& \underbrace{\frac{k}{n}[\log(1-\bar{\alpha}_{t})-1]}_{\text{diversity~$\uparrow$}}
+\underbrace{\frac{||\hat{\mu}-\mu||^{2}_{2}}{n}}_{\text{about bias~$\downarrow$}}
-\underbrace{\frac{\log|\hat\Sigma|}{n}+\frac{\mathrm{tr}(\hat{\Sigma})}{n(1-\bar{\alpha})}}_{\text{about variance~$\downarrow$}}
\\
&+\underbrace{\log\frac{1}{\delta}(\sup_{\mathcal{P}\in[0,1]} [\sum_{m=0}^{n}\mathrm{Bio}(m;n,\mathcal{P})\exp\{n\Delta(m/n,\mathcal{P})\}])\}}_{\text{about task} (\textit{i.e.} \text{,~the optimizee})}\text{.}
\end{align*}

\paragraph{Takeaways.} From the bound, we know that:
\begin{itemize}
    \item For any stochastic process at any time $t$, is a Gaussian distribution, the solution's dimension $k$ have to \textit{grow linearly} with the sample size $n$.
    \item A larger sample size $n$ reduces the generalization gap, \textit{i.e.}, sum of bias and variance. At a certain overall loss (\textit{e.g.}, the terminal phase of training), there is a classical bias-variance trade-off.
    \item The ability to generalize is also related to the kind of downstream task, with specific effects $\mathcal{M}$. As in the above example, $\mathcal{M}$ usually takes supremum for further concentration.
\end{itemize}

\paragraph{Theorem expansion.} Here we use the general distribution assumption for the stochastic process. Markov inequality in the proof can be replaced with different assumptions, \textit{e.g.}, using Hoeffding inequality for the sub-Gaussian, Bernstein inequality for the sub-exponential.

\paragraph{Theorem specialization.} Given different assumptions and tasks \textit{w.r.t.} $\mathcal{M}$ and $\Delta$, we have the Table~\ref{tab:mtd_tasks}. Previous works are related in order~\citep{langford2001bounds,mcallester1998some, alquier2018simpler}.
\begin{table}[h]
\tablestyle{2pt}{1.3}
    \centering
    \begin{tabular}{c}
         bound modifications \textit{w.r.t.} 
         $\Delta(a,b)$ on the left-hand side
         \\
         
         \shline
         $a\log\frac{a}{b}+(1-a)\log\frac{1-a}{1-b} \le \frac{1}{n}[\mathrm{KL}(q||p)++\log\frac{\sqrt{2n}}{\delta}]$
         \\
         $(b-a)^{2} \le{\frac{1}{2n}[\mathrm{KL(q||p)}+\log\frac{\sqrt{2n}}{\delta}]}$
         \\
         $b-a\le\frac{1}{\lambda}[\text{KL}(q||p)-\log(\delta)+\frac{\lambda}{n}(b-a)]$
    \end{tabular}
    \caption{Specialization: varied distance function $\Delta$.}
    \label{tab:mtd_tasks}
    \vspace{-10pt}
\end{table}

\subsection{Add-on: optimal-free and dimension-free}
\texttt{oracle} is a neural network to generate initial points. It remembers the suboptimal solutions. Training from scratch is avoided. More over, an element-wise variant for dynamic dimension $k=\dim x$ is provided in the Appendix.
\begin{algorithm}
\caption{Alternative \texttt{oracle}: optimal generator}
\begin{algorithmic}
\FOR{given \#epochs}
    \STATE ${x}_0 \gets \texttt{oracle}({g})$
    \STATE $\mathcal{L}_\text{pre} \gets f({\theta}, {x}_0)$
    \STATE Update \texttt{oracle} by minimizing $\mathcal{L}_\text{pre}$
    \STATE Generate the forward trajectory starting from ${x}_0$: $\{\tilde{x}_0, \tilde{x}_1, \dots, \tilde{x}_\mathtt{T}\}$
    \STATE Train \texttt{opt} using Algorithm~\ref{alg:mtd_diff_diffl2o_training} for one epoch 
    \STATE $\mathcal{L}_\text{post} = \text{MSE}({x}_0, {\hat{x}}_0)$\\
    \STATE Update \texttt{oracle} by minimizing $\mathcal{L}_\text{post}$
\ENDFOR
\end{algorithmic}
\end{algorithm}
\vspace{-20pt}

\section{Empirical Evaluation}
\subsection{Overview}
Numerical evaluations are built on conventional optimization problems, including convex and non-convex cases. Diff-L2O is applicable on the parameter solution space of the neural network. Summary: 1) vanilla Diff-L2O works very well on non-convex problems; 2) Diff-L2O brings a huge enhancement to the conventional optimizers; 3)

\subsection{Settings}
\paragraph{Compared baselines.} 
We compare various analytical optimizers (Gradient Descent and Adam~\citep{kingma2014adam}) and learned optimizers (L2O-DM~\citep{andrychowicz2016learning} and L2O-RNNProp~\citep{lv2017learning}). For learned optimizers, we train them on the same set of samples.

\paragraph{Training hyperparameters.} The maximum step $\mathtt{T}$ is set to 100 when training \texttt{opt}. \#Diffusion steps for inference is 100. The coefficient for variance scheduling range from $1\times 10^{-5}$ to $2\times 10^{-2}$, linearly increasing  along $t$. The coefficient $\gamma$ for loss balancing is set to $0.5$ as default. 

\paragraph{Optimizees' hyperparameters.} Diff-L2O is evaluated on three representative optimization problems with varied complexities and characteristics. For all optimizees, training and testing samples are independently drawn from a standard Gaussian distribution $\mathcal{N}(0,\mathbf{I})$. For example, in LASSO, $\mathbf{A}$ and $\mathbf{b}$ are sampled from standard Gaussian, simplified as $\theta$.
\begin{equation}
    \boldsymbol{x}^{\mathrm{LASSO}} = \mathop{\arg\min}_{\boldsymbol{x}} \ \frac{1}{2}\|\mathbf{A}\boldsymbol{x}-\mathbf{b} \|_2^2 + \lambda \|\boldsymbol{x}\|_1
\end{equation}
Other formulations of classic problems about Rastrigin and Ackley are in Appendix.

$\rhd$ \textit{LASSO} Two problem scales is related: a low-dimensional setting with design matrix $\mathbf{A} \in \mathbb{R}^{5 \times 10}$ and a medium-dimensional setting with $\mathbf{A} \in \mathbb{R}^{25 \times 50}$. The $\ell_1$ regularization coefficient is fixed at $\lambda = 0.005$ for both configurations.

$\rhd$ \textit{Rastrigin} We investigate both low-dimensional ($d=2$) and high-dimensional ($d=10$) scenarios. The amplitude of the modulation term is set to $\alpha=10$, which controls the intensity of local minima. It's non-convex.

$\rhd$ \textit{Ackley} Similar to the Rastrigin function, we examine the optimization performance in both low-dimensional ($d=2$) and high-dimensional ($d=10$) spaces. It's non-convex.

\begin{figure}[t]
  \begin{center}
    \begin{subfigure}
      \centering      \includegraphics[width=0.46\linewidth]{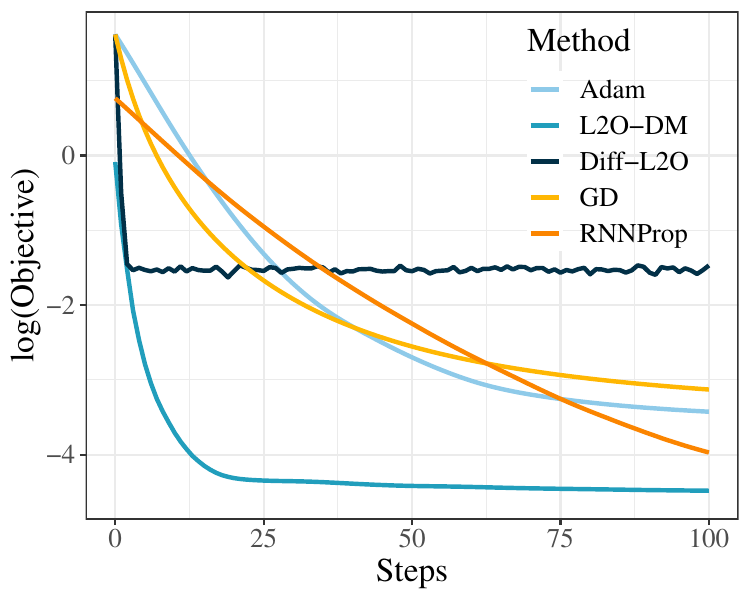}
    \label{fig:comparison_lasso_5}
    \end{subfigure}
    \quad
    \begin{subfigure}
      \centering      \includegraphics[width=0.46\linewidth]{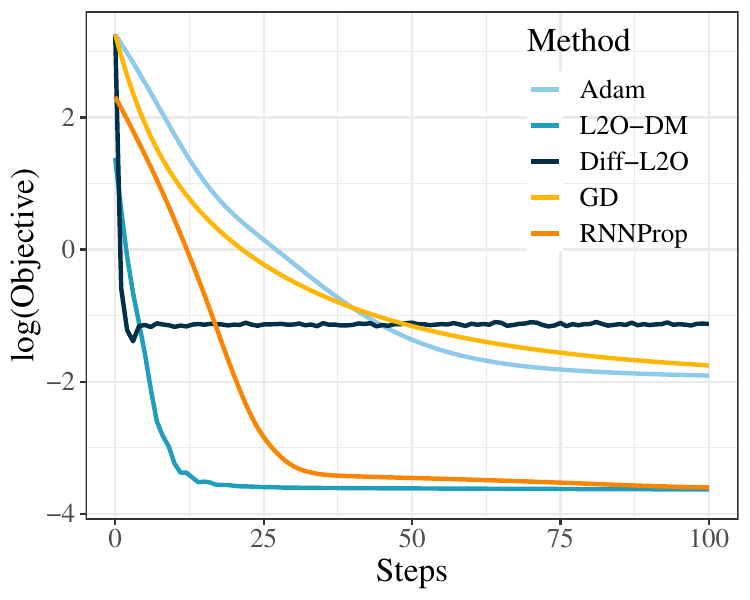}
      \label{fig:comparison_lasso_25}
    \end{subfigure}
  \end{center}
  \vspace{-20pt}
  \begin{center}
      LASSO $\dim_{x}=10$ (left) and $\dim_{x}=50$ (Right).
  \end{center}
    \begin{center}
    \begin{subfigure}
      \centering      \includegraphics[width=0.46\linewidth]{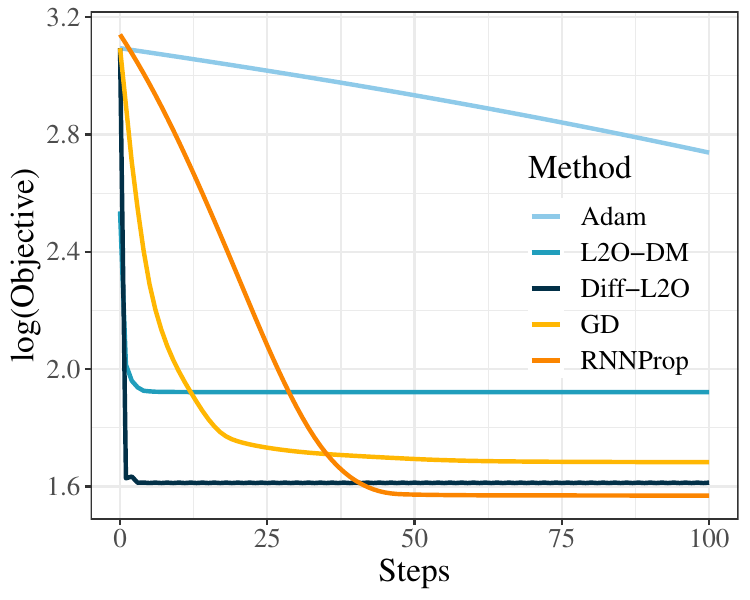}
      \label{fig:comparison_ras_2}
    \end{subfigure}
    \quad
    \begin{subfigure}
      \centering      \includegraphics[width=0.46\linewidth]{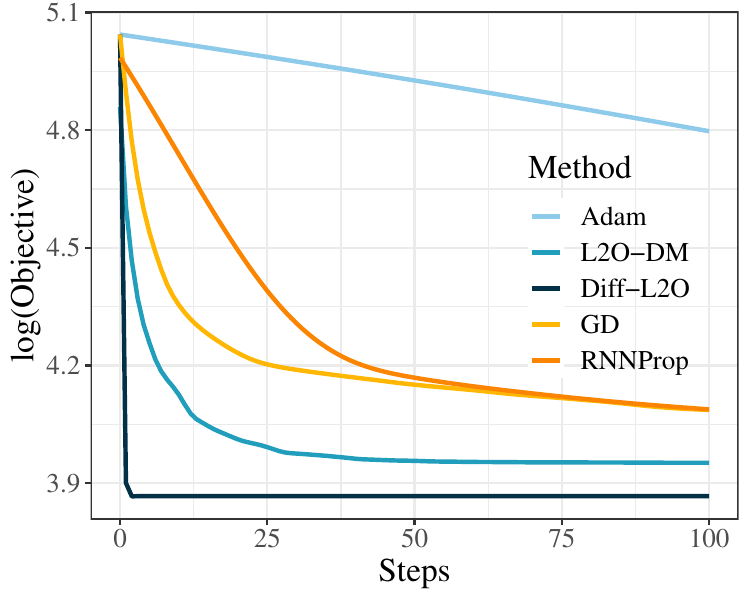}
      \label{fig:comparison_ras_10}
    \end{subfigure}
  \end{center}
  \vspace{-20pt}
  \begin{center}
      Rastrigin $\dim_{x}=2$ (left) and $\dim_{x}=10$ (Right).
  \end{center}
  \begin{center}
    \begin{subfigure}
      \centering      \includegraphics[width=0.46\linewidth]{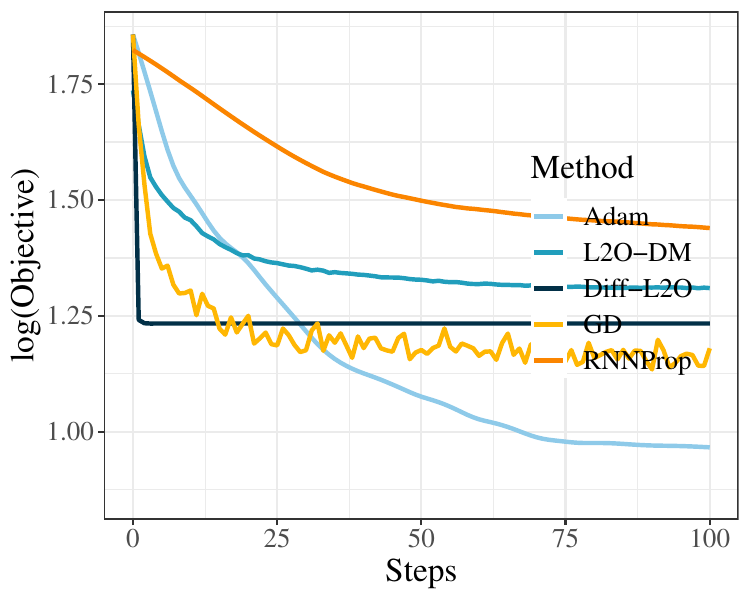}\label{fig:comparison_ackley_2}
    \end{subfigure}
    \quad
    \begin{subfigure}
      \centering      \includegraphics[width=0.46\linewidth]{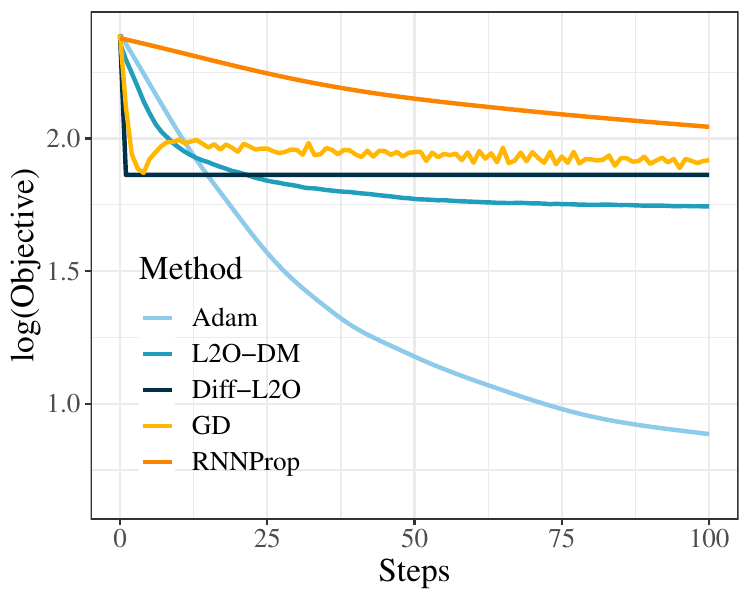}\label{fig:comparison_ackley_10}
    \end{subfigure}
  \end{center}
  \vspace{-15pt}
  \begin{center}
      Ackley $\dim_{x}=2$ (left) and $\dim_{x}=10$ (Right).
  \end{center}
  \caption{Comparison on optimizees across \#dimension: LASSO, Rastrigin and Ackley.}  \label{fig:comparison_optimization}
\end{figure}
\begin{figure}[t]
    \centering
    \begin{center}
    \begin{subfigure}
      \centering      \includegraphics[width=0.46\linewidth]{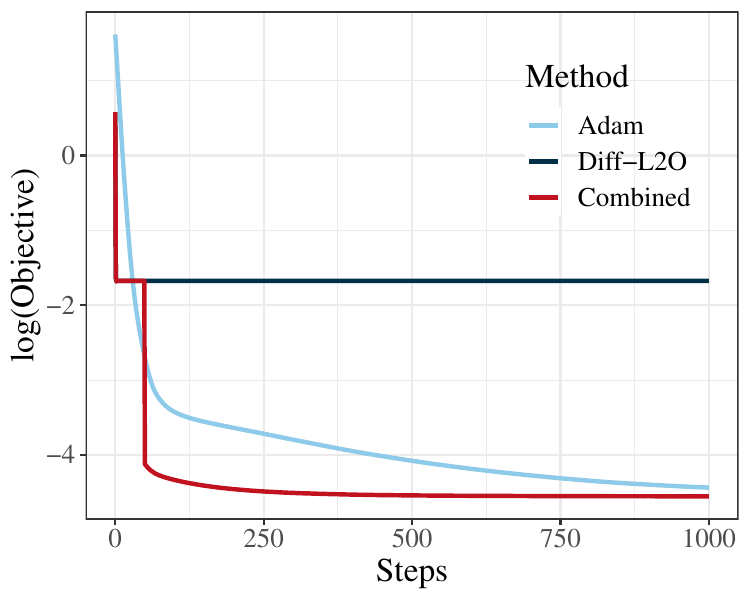}
    \label{fig:comparison_lasso_5}
    \end{subfigure}
    \quad
    \begin{subfigure}
      \centering      \includegraphics[width=0.46\linewidth]{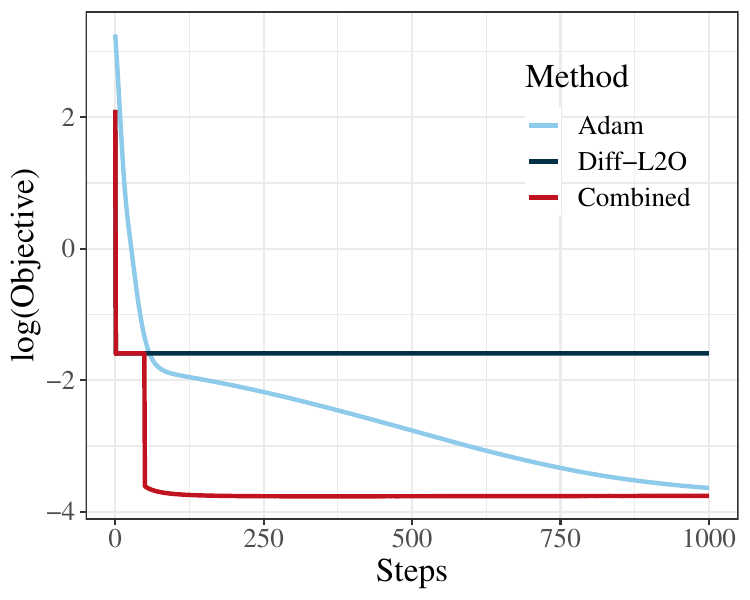}
      \label{fig:comparison_lasso_25}
    \end{subfigure}
  \end{center}
  \vspace{-20pt}
  \begin{center}
      LASSO $\dim_{x}=10$ (left) and $\dim_{x}=50$ (Right).
  \end{center}
    \begin{center}
    \begin{subfigure}
      \centering      \includegraphics[width=0.46\linewidth]{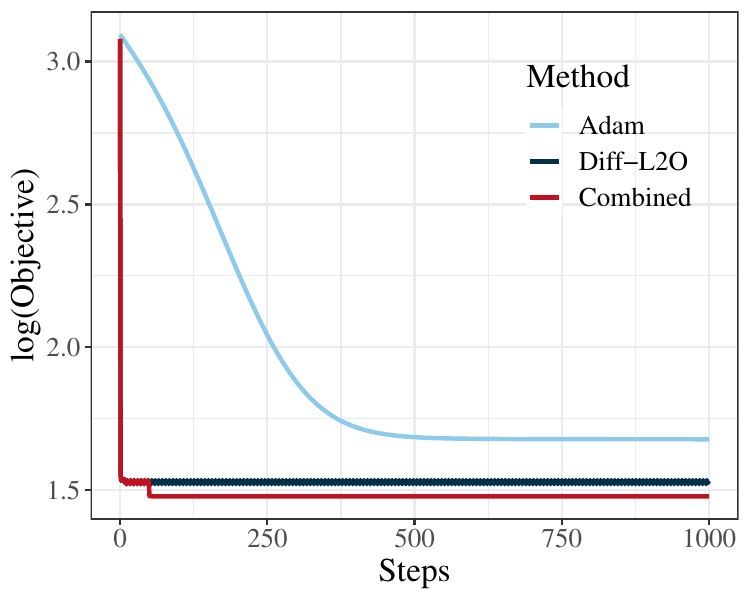}
      \label{fig:comparison_ras_2}
    \end{subfigure}
    \quad
    \begin{subfigure}
      \centering      \includegraphics[width=0.46\linewidth]{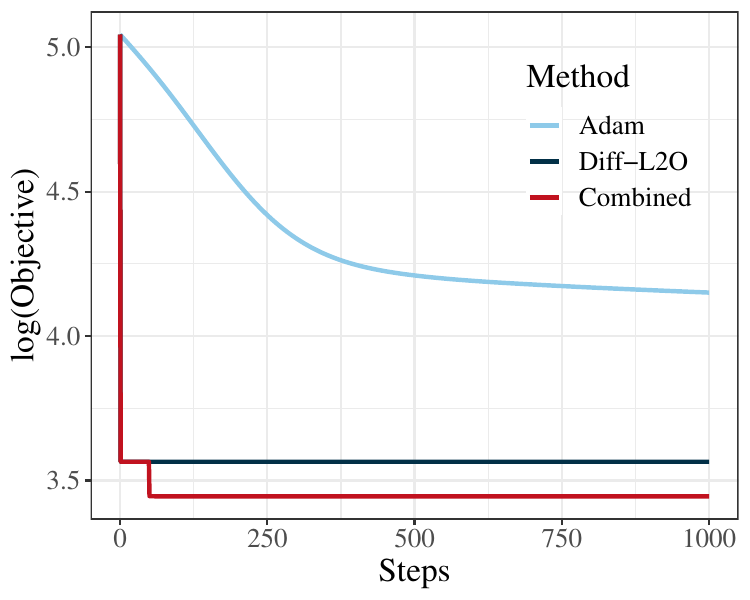}
      \label{fig:comparison_ras_10}
    \end{subfigure}
  \end{center}
  \vspace{-20pt}
  \begin{center}
      Rastrigin $\dim_{x}=2$ (left) and $\dim_{x}=10$ (Right).
  \end{center}
  \begin{center}
    \begin{subfigure}
      \centering      \includegraphics[width=0.46\linewidth]{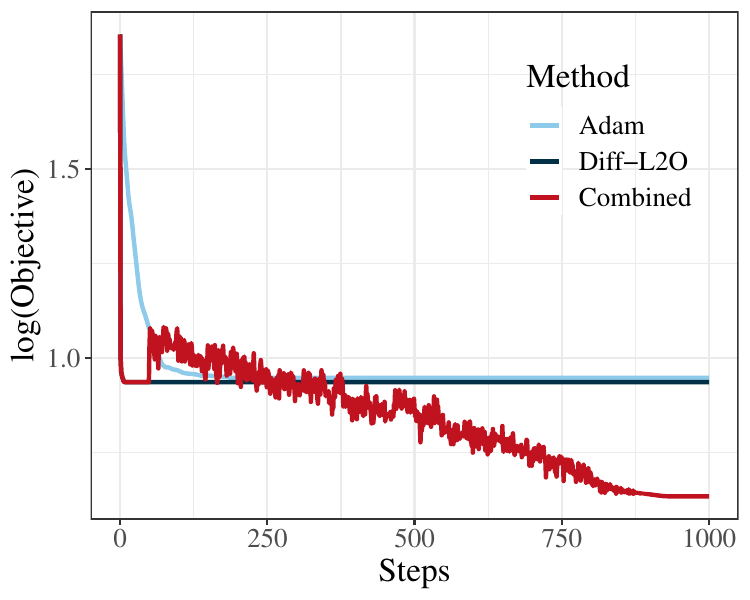}\label{fig:comparison_ackley_2}
    \end{subfigure}
    \quad
    \begin{subfigure}
      \centering      \includegraphics[width=0.46\linewidth]{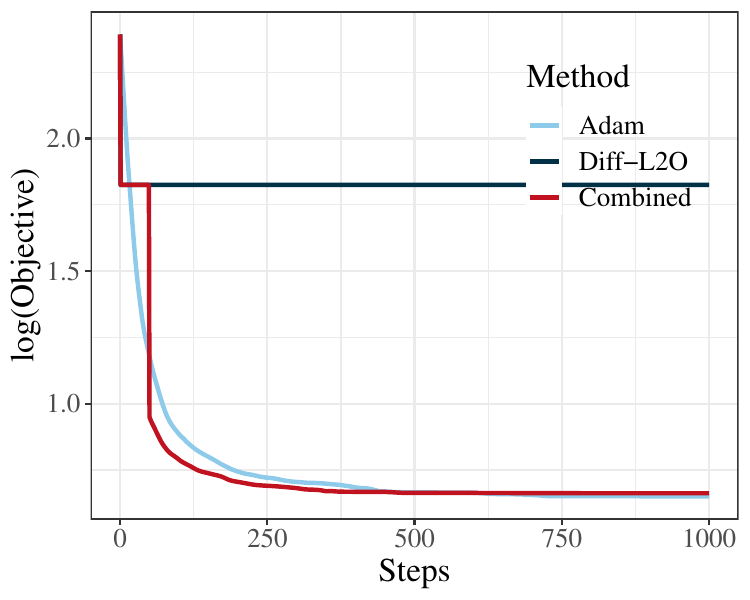}\label{fig:comparison_ackley_10}
    \end{subfigure}
  \end{center}
  \vspace{-15pt}
  \begin{center}
  Ackley $\dim_{x}=2$ (left) and $\dim_{x}=10$ (Right). \end{center}
    \caption{Ablation: compatibility of Diff-L2O with conventional optimizers.}
    \label{fig:abl_compatibility}
\end{figure}
\subsection{Comparison}
\paragraph{LASSO.} We first conduct experiments on the LASSO optimizees and compare the performance on unseen optimizee problems. The experimental results are summarized in Figure~\ref{fig:comparison_optimization}. We can observe that \methodname has fast convergence speed compared to other baselines, achieving near-convergence range with less than ten steps. In the absence of gradient information, {Diff-L2O} converges to the wall of the LASSO convex valley. This issue can be easily resolved by combining {Diff-L2O} and analytical optimizers to achieve more accurate solutions.

\paragraph{Rastrigin.} In Rastrigin tasks, our method has demonstrated faster convergence speed and also similar or higher quality compared to baselines. Specifically, \methodname achieves a loss objective of 44.09 within 10 steps, while the most competitive baseline, \textit{i.e.} RNNProp, can only achieve a loss of 56.68 in 100 steps. Such an advantage is enlarged in higher dimension of the variables as the benchmark method suffers from the curse of dimensionality, while our method perform consistently for different dimensions. 

\paragraph{Ackley.} On the Ackley tasks, \methodname also out-performs existing baseline methods with clear margins: In 2-dimensional case, \methodname achieves a loss objective of 3.15 within 10 steps, comparing to the most competitive baseline, \textit{i.e.} RNNProp, which can only achieve a loss of 4.48 in 10 steps. In 10-dimensional case, \methodname achieves a loss objective of 5.37 within 10 steps, while the most competitive baseline can only achieve a loss of 6.08 in 100 steps. Analytical optimizers such as Adam out-performs all L2O methods due to the moderate difficulty of Ackley problems.

\paragraph{MNIST on DNN.} We evaluate the classification 
performance of {Diff-L2O} on MNIST. In Figure~\ref{fig:comparison_mnist}, and it achieved a loss of 0.228 and accuracy of 92.06\% on test set, which out-perform the RNNProp method that achieve a loss of 0.268 and accuracy of 90.28, and L2O-DM with a loss of 0.252 and accuracy of 90.79 on the same test set. Detailed settings on DNN and loss curves are placed in Appendix.

\subsection{Ablation}
\paragraph{Ablation: compatibility with conventional optimizers.} %
\textit{Diff-L2O works well when adapted to other methods.}
The stochastic nature of diffusion models enables rapid initial convergence but may slow in later stages, which is particularly disadvantageous for convex problems. This motivates our hybrid approach: using the diffusion model for initialization followed by traditional optimizers. Our results show this strategy consistently outperforms conventional optimizers on both convex and non-convex problems.

\paragraph{Settings.} %
We evaluate all optimizees on the same test set as the comparison experiments. Our hybrid optimization consists of two phases: an initial exploration phase utilizing our diffusion-based model for the first 50 iterations, followed by a fine-grained fine-tuning phase with the Adam optimizer. 

\paragraph{Analyses.}
Fig.~\ref{fig:comparison_optimization} and~\ref{fig:abl_compatibility} show that, in the comparison experiment' convex case, the performance using a vanilla Diff-L2O can be improved by using a combination of conventional optimizers. Diff-L2O can be used to quickly generate foundational solutions with a small amount of fine-tuning to reach the optimal.

\paragraph{Ablation: optimal-free.} %
The training of diffusion models requires solving numerous optimization problems of the same optimizee family, which inherently limits the model's generalizability. The \texttt{oracle} component offers a potential solution to this limitation. Therefore, we conduct an ablation study to analyze how different oracle configurations impact the model's performance.

\paragraph{Settings.} %
We conduct a series of experiments to understand the effects of introduced components: (1) \textit{Noisy}: where we replace \texttt{oracle} with a module that  generates random noises; (2) \textit{Fixed}: where we do not update the \texttt{oracle} network; (3) \textit{Partial}, where we update the \texttt{oracle} network with $\mathcal{L}_\text{pre}$ only; and (4) \textit{Perfect}: where we make \texttt{oracle} to always output the optimal solutions.

\paragraph{Analyses.} According to Tab.~\ref{tab:exp_abl_oracle}, In case: noisy, we find that random initialization with poor performance. It show us that initialization strategy is necessary, even a fixed pre-trained network. Loss term $\mathcal{L}_{\text{pre}}$, lowering task loss, helps by making better initial points. The benefits, however, are increased gradually comparing to perfect cases. Loss term $\mathcal{L}_{\text{post}}$, closing backward and forward processes, shows the importance of samples with great diversity. All these modules lead DIff-L2O's performance closer to the perfect cases (starting at the optimal).

\begin{table}[t]
    \centering
    \tablestyle{12pt}{1.3}
    \caption{Log loss with different variants of oracles.}
    \vspace{-2mm}
    \label{tab:exp_abl_oracle}
    \begin{tabular}{cccc}
        variants & LASSO & Rastrigin & Ackley \\
        \shline noisy & -1.306 & 1.727 & 1.301\\
        fixed & -1.427 & 1.657 & 1.281\\
        partial & -1.456 & 1.627 & 1.257\\
        perfect & -1.676 & 1.532 & 0.936\\
        \default{Ours} & \default{-1.660} & \default{1.601} & \default{1.233}\\
    \end{tabular}
    
\end{table}

\begin{table}[t]
    \centering    \tablestyle{8pt}{1.3}

    \caption{Log loss with different guidance vector. }
    \label{tab:exp_abl_guidance}
    \begin{tabular}{ccccc}
        \multirow{2}{*}{variants} & LASSO & LASSO &Rastrigin& Rastrigin\\
                 & (t=10) & (t=100) &(t=10) & (t=100)\\
        \midrule gradient & -3.161& -4.011 & 3.064 & 2.738\\
        global & -1.674 & -1.673 & 1.532 & 1.532\\
        all & -3.153 & -3.938 & 1.618 & 1.643\\
    \end{tabular}
    
\end{table}

\paragraph{Ablation: guidance.} %
In this part, the guidance vector ${g}$ can be time step $t$ dependent, and we denote it by ${g}_t$. In practice, ${g}_t$ is a crucial component in Diff-L2O. For convex problems like LASSO, incorporating gradient information in the guidance vector can significantly improve the convergence speed and accuracy. However, in non-convex problems such as Rastrigin, the gradient can potentially be a source of noise that guides the solutions to local minima.

\paragraph{Settings.} %
we conducted experiments on LASSO and Rastrigin optimizees using three types of guidance vectors: (1) \textit{Gradient}, where only the gradient is considered as the guidance vector; (2) \textit{Global}, where the optimizees' parameters $\mathbf{\theta}$ are used as the guidance vector; and (3) \textit{All}, where the guidance vector consists of both gradient and $\mathbf{\theta}$.

\paragraph{Analyses.} 
In 1) convex cases, as shown in Tbl.~\ref{tab:exp_abl_guidance}, the gradient largely guides whether the current point is optimal or not and contains useful local information. The cases, with gradient only, are dominated by the first-order information, and thus got a log loss value of -3.161 and -4.011 from -3.153 and -3.938. 2) The information about the solutions, in non-convex cases, are not determined by the gradient. However, a lot of samples are still helpful, as the results 1.618 from $t=10$ converged quickly comparing to 1.643 from $t=100$, and the sampling have not yet converged in Gradient case with a gap of $0.326$.

\paragraph{Evaluation: training time.} %
Table~\ref{tab:tbl_time_cost} demonstrates the training time of L2O-DM~\citep{andrychowicz2016learning} and our method. It can be clearly seen that the \methodname can be trained rapidly, using merely $2\%$ of time compared to L2O algorithms. This rapid training makes our model practical. 

\paragraph{Settings.} %
The experiments are conducted in default settings on GPU 1$\times$NVIDIA-A100 and CPU AMD EPYC 7H12 64-Core. \#iterations is 100.

\begin{table}[t]
    \centering    \tablestyle{11pt}{1.3}

    \caption{Training time of L2O-DM and \methodnameE. }
    \vspace{+0mm}
    \label{tab:tbl_time_cost}
    \begin{tabular}{ccc}
        optimizees & L2O-DM & \default{\methodname} \\
    \shline
        LASSO (5-dim) & $\sim4$ hours & \default{203 s}\\
        LASSO (25-dim) & $\sim6$ hours & \default{376 s}\\
        Rastrigin (2-dim) & $\sim2$ hours & \default{310 s}\\
        Rastrigin (10-dim) &  $\sim2$ hours & \default{393 s}\\
        Ackley (2-dim) &  $\sim3$ hours & \default{309 s}\\
        Ackley (10-dim) &  $\sim3$ hours & \default{543 s}\\
    \end{tabular}
    \vspace{-2mm}
\end{table}

\begin{figure}
    \centering
    \includegraphics[width=0.6\linewidth]{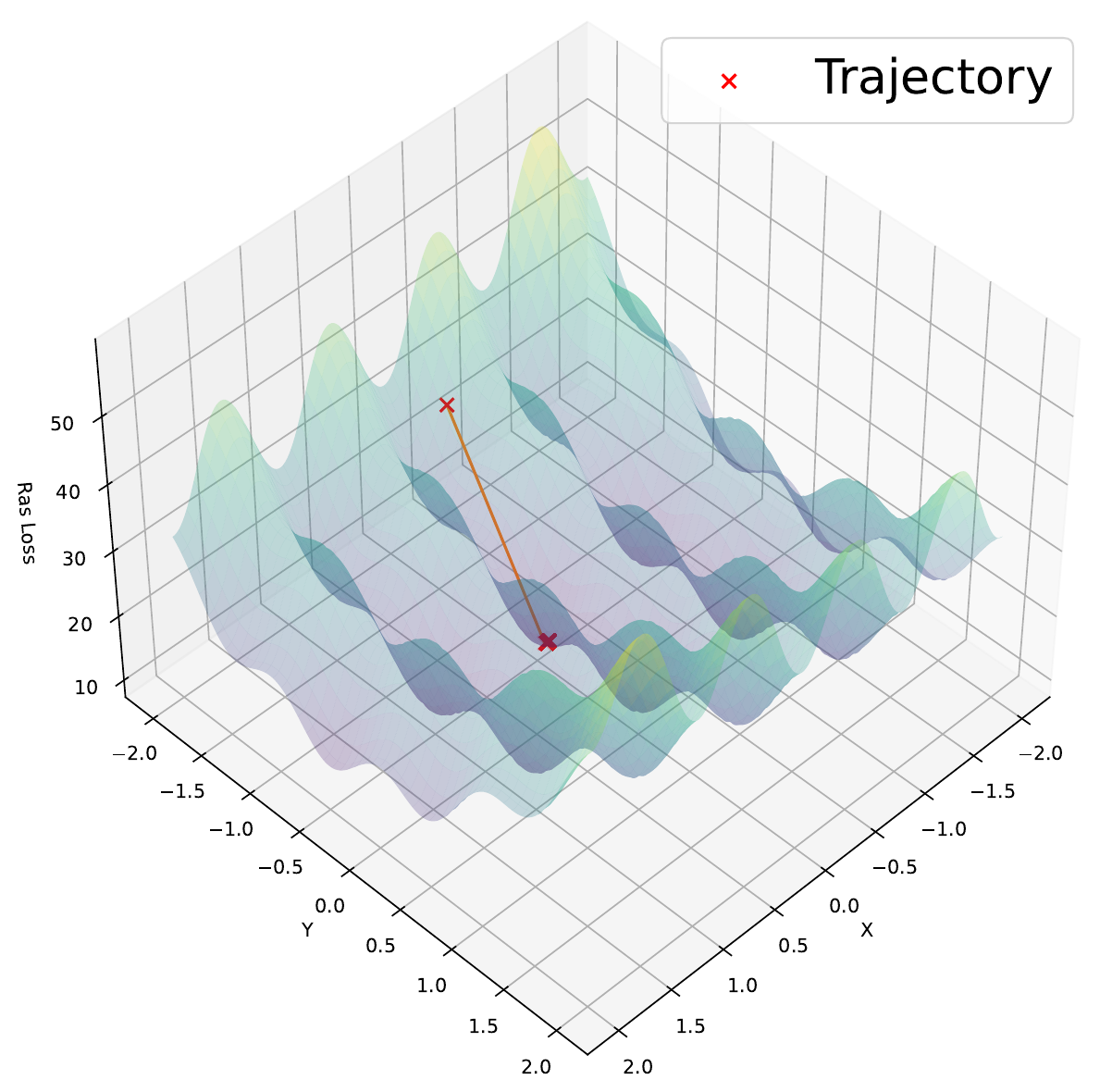}
    \caption{Visualization: learning surface. Fast convergence happens within several epochs.}
    \label{fig:vsl_ras}
\end{figure}
\paragraph{Visualization: trajectories.} %
We demonstrate that Diff-L2O rapidly approaches the vicinity of optimal solutions in the early stages, notably within the first iteration.

\paragraph{Settings.} %
We set the dimension for all optimizees (LASSO, Rstrigin, Ackley) to $2$ with other hyperparameters the same.

\paragraph{Analyses.}
Even in the non-convex case, Rastrigin, the learned descent trajectory of the optimizer reaches the area around the global optimum in almost the starting iterations.

\paragraph{Visualization: modeled distribution.} %
\paragraph{Settings.} %
The dimension of all optimizees are set to $2$ and other hyperparameters keep unchanged. The learned and true distributions mean Diff-L2O in default setting and gradient descent, respectively, with 5000 initial points.

\paragraph{Analyses.} The learned distribution and the distribution gotten from conventional optimizer are matched generally. The diversity of learned distribution are greater.
\begin{figure}[htbp]
    \centering
   \includegraphics[scale=0.625]{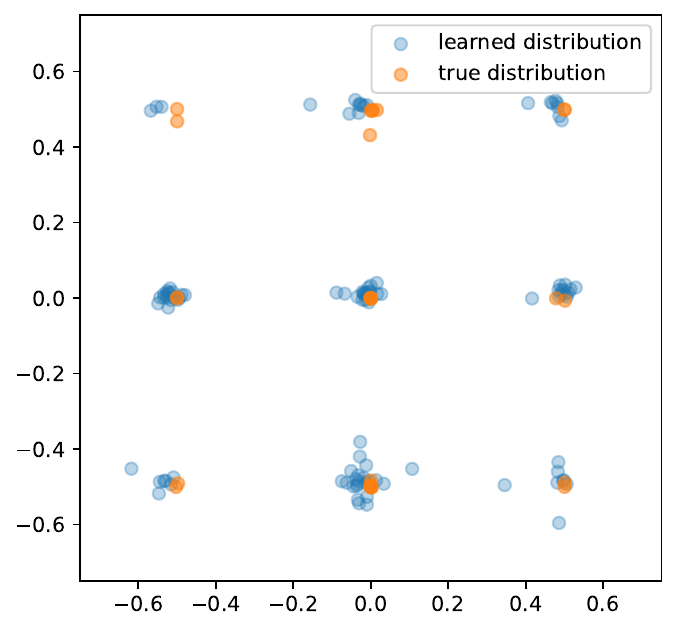}
   \vspace{-1.5mm}
   \caption{Visualization of the learned and the ground-truth distribution (true). The distributions are generally matched.}
   \label{fig:exp_vsl_distribution}
   \vspace{-5.8mm}
\end{figure}
\section{Related Works}
\paragraph{Learning to optimize (L2O).} L2O is an alternative optimization paradigm that aims to learn effective optimization rules in a data-driven way. It generates optimization rules based on the performance on a set of training problems. it has demonstrated success on a wide range of tasks, including black-box optimization~\citep{chen2017learning,krishnamoorthy2023diffusion}, Bayesian optimization~\citep{cao2019learning}, minimax optimization~\citep{shen2021learning,jiang2018learning} and domain adaptation~\citep{chen2020automated,li2020halo}. More recently, L2O has demonstrated its ability of solving large-scale problems~\citep{metz2022velo,chen2022scalable}, making it more practical for broader applications, \textit{e.g.}, conditional generation~\citep{wang2024neural, wang2025recurrent}.  

The architectures of the learnable optimizer for L2O works have undergone great evaluation. 
In the seminal work of \citet{andrychowicz2016learning}, a coordinate-wise long-short-term memory (LSTM) network~\cite{hochreiter1997long} is adopted as the backbone, which can capture the inter-parameter dependencies with low computational overhead. Subsequently, while some works~\citep{vicol2021unbiased} have utilized multi-layer perceptions (MLPs) for learnable optimization, a large portion of L2O works have adopted the recurrent neural networks (RNNs)~\cite{rumelhart1986learning} as the architecture of their learnable optimizer~\citep{chen2021learning}. For example, \citet{shen2021learning} proposes to use two LSTM networks to solve min-max optimization problems. \citet{cao2019learning} deploys multiple LSTM networks to tackle population-based problems.  %
Later on, researchers have explored the possibility of using Transformers~\citep{vaswani2017attention} as learnable optimizers. \citet{chen2022towards} proposes to use Transformer as a tool for hyperparameter optimization. \citet{jain2023mnemosyne,gartner2023transformer} propose L2O frameworks that apply Transformers to solve general optimization problem and achieves faster convergence compared to traditional algorithms such as SGD and Adam~\citep{kingma2014adam}. 
In this paper, we propose to apply a different paradigm, \textit{i.e.,} diffusion, as the foundation of our L2O framework. This framework model solution space with a fine-grained approximation. 

\paragraph{Diffusion models.} Diffusion probabilistic models~\cite{ho2020denoising,song2020denoising} have emerged as a powerful tool for generating high-quality samples with different modalities such as images~\citep{dhariwal2021diffusion,rombach2022high}, texts~\cite{gong2022diffuseq,xu2023versatile}, 3d objects~\cite{erkocc2023hyperdiffusion,gu2023learning}, and videos~\citep{ho2022imagen}. These models have demonstrated on-par or better generation quality compared to their precursors such as generative adversarial networks (GANs)~\citep{goodfellow2020generative,odena2017conditional,gong2019twin}. In a typical training pipeline, diffusion models learn their parameters through iterative addition and removal of noises; and in the inference stage, they begin with a randomly sampled noise and generate the corresponding sample by iteratively denoising. Conditional diffusion models as an important branch of diffusion models, such as those in~\cite{ho2022classifier,liu2022compositional,chao2022denoising}, enables generations with clear instruction. In this study, we introduce a novel conditional diffusion model that operates within the solution space of optimization problems including weight of neural networks. Empirically, diffusion models work well.
\section{Conclusion}
This work propose a novel L2O framework Diff-L2O. It use diffusion model to model solution space, accelerating the optimization process. We discuss the key to modeling the solution space while giving related generalization bound. Diff-L2O is empirically verified to achieve significant results on multiple benchmarks, which further validates our analyses and discussion.

Furthermore, the ablation study reveals the essence of designed components in Diff-L2O, and the combined method demonstrate huge potential for implementing our method as initialization in practice, which is especially useful when analytical properties are essential.

\newpage

\section*{Impact Statement}
This paper aims to improve the field of learning-to-optimize (L2O) by introducing of \methodnameE, a novel framework based on diffusion modeling. it has the potential to impact various fields that rely on optimization techniques, such as machine learning, operations research, and engineering. By efficiently generating high-quality solutions for diverse optimization problems, \methodname can contribute to the development of more effective and efficient optimization algorithms that enable researchers and practitioners to solve complex problems. Our work has several potential societal implications which are general in this field, but we don’t feel any of them require special attention at this time.

\bibliography{main_icml2025}

\begin{thebibliography}{55}
\providecommand{\natexlab}[1]{#1}
\providecommand{\url}[1]{\texttt{#1}}
\expandafter\ifx\csname urlstyle\endcsname\relax
  \providecommand{\doi}[1]{doi: #1}\else
  \providecommand{\doi}{doi: \begingroup \urlstyle{rm}\Url}\fi

\bibitem[Alquier \& Guedj(2018)Alquier and Guedj]{alquier2018simpler}
Alquier, P. and Guedj, B.
\newblock Simpler pac-bayesian bounds for hostile data.
\newblock \emph{Machine Learning}, 107\penalty0 (5):\penalty0 887--902, 2018.

\bibitem[Andrychowicz et~al.(2016)Andrychowicz, Denil, Colmenarejo, Hoffman, Pfau, Schaul, Shillingford, and De~Freitas]{andrychowicz2016learning}
Andrychowicz, M., Denil, M., Colmenarejo, S.~G., Hoffman, M.~W., Pfau, D., Schaul, T., Shillingford, B., and De~Freitas, N.
\newblock Learning to learn by gradient descent by gradient descent.
\newblock In \emph{Advances in Neural Information Processing Systems (NIPS)}, volume~29, pp.\  3981--3989, 2016.

\bibitem[Attouch et~al.(2019)Attouch, Chbani, Fadili, and Riahi]{attouch2019first}
Attouch, H., Chbani, Z., Fadili, J., and Riahi, H.
\newblock First-order optimization algorithms via inertial systems with hessian driven damping.
\newblock \emph{arXiv preprint}, arXiv:1907.10536, 2019.

\bibitem[Bowman \& Mont{\'u}far(2022)Bowman and Mont{\'u}far]{bowman2022implicit}
Bowman, B. and Mont{\'u}far, G.
\newblock Implicit bias of mse gradient optimization in underparameterized neural networks.
\newblock \emph{arXiv preprint arXiv:2201.04738}, 2022.

\bibitem[Cao et~al.(2019)Cao, Chen, Wang, and Shen]{cao2019learning}
Cao, Y., Chen, T., Wang, Z., and Shen, Y.
\newblock Learning to optimize in swarms.
\newblock In Wallach, H., Larochelle, H., Beygelzimer, A., d\textquotesingle Alch\'{e}-Buc, F., Fox, E., and Garnett, R. (eds.), \emph{Advances in Neural Information Processing Systems}, volume~32. Curran Associates, Inc., 2019.

\bibitem[Chao et~al.(2022)Chao, Sun, Cheng, Lo, Chang, Liu, Chang, Chen, and Lee]{chao2022denoising}
Chao, C.-H., Sun, W.-F., Cheng, B.-W., Lo, Y.-C., Chang, C.-C., Liu, Y.-L., Chang, Y.-L., Chen, C.-P., and Lee, C.-Y.
\newblock Denoising likelihood score matching for conditional score-based data generation.
\newblock \emph{arXiv preprint arXiv:2203.14206}, 2022.

\bibitem[Chen et~al.(2021)Chen, Chen, Chen, Heaton, Liu, Wang, and Yin]{chen2021learning}
Chen, T., Chen, X., Chen, W., Heaton, H., Liu, J., Wang, Z., and Yin, W.
\newblock Learning to optimize: A primer and a benchmark.
\newblock \emph{arXiv preprint arXiv:2103.12828}, 2021.

\bibitem[Chen et~al.(2022{\natexlab{a}})Chen, Chen, Chen, Heaton, Liu, Yin, and Wang]{chen2022learning}
Chen, T., Chen, X., Chen, W., Heaton, H., Liu, J., Yin, W., and Wang, Z.
\newblock Learning to optimize: A primer and a benchmark.
\newblock \emph{Journal of Machine Learning Research}, 23:\penalty0 1--59, 2022{\natexlab{a}}.

\bibitem[Chen et~al.(2020)Chen, Yu, Wang, and Anandkumar]{chen2020automated}
Chen, W., Yu, Z., Wang, Z., and Anandkumar, A.
\newblock Automated synthetic-to-real generalization.
\newblock In \emph{International Conference on Machine Learning (ICML)}, pp.\  1746--1756, 2020.

\bibitem[Chen et~al.(2022{\natexlab{b}})Chen, Chen, Cheng, Chen, Awadallah, and Wang]{chen2022scalable}
Chen, X., Chen, T., Cheng, Y., Chen, W., Awadallah, A., and Wang, Z.
\newblock Scalable learning to optimize: A learned optimizer can train big models.
\newblock In \emph{European Conference on Computer Vision}, pp.\  389--405. Springer, 2022{\natexlab{b}}.

\bibitem[Chen et~al.(2017)Chen, Hoffman, Colmenarejo, Denil, Lillicrap, Botvinick, and De~Freitas]{chen2017learning}
Chen, Y., Hoffman, M.~W., Colmenarejo, S.~G., Denil, M., Lillicrap, T.~P., Botvinick, M., and De~Freitas, N.
\newblock Learning to learn without gradient descent by gradient descent.
\newblock In \emph{International Conference on Machine Learning (ICML)}, pp.\  748--756, 2017.

\bibitem[Chen et~al.(2022{\natexlab{c}})Chen, Song, Lee, Wang, Zhang, Dohan, Kawakami, Kochanski, Doucet, Ranzato, et~al.]{chen2022towards}
Chen, Y., Song, X., Lee, C., Wang, Z., Zhang, R., Dohan, D., Kawakami, K., Kochanski, G., Doucet, A., Ranzato, M., et~al.
\newblock Towards learning universal hyperparameter optimizers with transformers.
\newblock \emph{Advances in Neural Information Processing Systems}, 35:\penalty0 32053--32068, 2022{\natexlab{c}}.

\bibitem[Dauber et~al.(2020)Dauber, Feder, Koren, and Livni]{dauber2020can}
Dauber, A., Feder, M., Koren, T., and Livni, R.
\newblock Can implicit bias explain generalization? stochastic convex optimization as a case study.
\newblock \emph{Advances in Neural Information Processing Systems}, 33:\penalty0 7743--7753, 2020.

\bibitem[Dhariwal \& Nichol(2021)Dhariwal and Nichol]{dhariwal2021diffusion}
Dhariwal, P. and Nichol, A.
\newblock Diffusion models beat gans on image synthesis.
\newblock \emph{Advances in neural information processing systems}, 34:\penalty0 8780--8794, 2021.

\bibitem[Erko{\c{c}} et~al.(2023)Erko{\c{c}}, Ma, Shan, Nie{\ss}ner, and Dai]{erkocc2023hyperdiffusion}
Erko{\c{c}}, Z., Ma, F., Shan, Q., Nie{\ss}ner, M., and Dai, A.
\newblock Hyperdiffusion: Generating implicit neural fields with weight-space diffusion.
\newblock \emph{arXiv preprint arXiv:2303.17015}, 2023.

\bibitem[Garipov et~al.(2018)Garipov, Izmailov, Podoprikhin, Vetrov, and Wilson]{garipov2018loss}
Garipov, T., Izmailov, P., Podoprikhin, D., Vetrov, D.~P., and Wilson, A.~G.
\newblock Loss surfaces, mode connectivity, and fast ensembling of dnns.
\newblock \emph{Advances in neural information processing systems}, 31, 2018.

\bibitem[Gartner et~al.(2023)Gartner, Metz, Andriluka, Freeman, and Sminchisescu]{gartner2023optimus}
Gartner, E., Metz, L., Andriluka, M., Freeman, C.~D., and Sminchisescu, C.
\newblock Transformer-based learned optimization.
\newblock In \emph{Proceedings of the IEEE/CVF Conference on Computer Vision and Pattern Recognition (CVPR)}, 2023.

\bibitem[G{\"a}rtner et~al.(2023)G{\"a}rtner, Metz, Andriluka, Freeman, and Sminchisescu]{gartner2023transformer}
G{\"a}rtner, E., Metz, L., Andriluka, M., Freeman, C.~D., and Sminchisescu, C.
\newblock Transformer-based learned optimization.
\newblock In \emph{Proceedings of the IEEE/CVF Conference on Computer Vision and Pattern Recognition}, pp.\  11970--11979, 2023.

\bibitem[Gong et~al.(2019)Gong, Xu, Li, Zhang, and Batmanghelich]{gong2019twin}
Gong, M., Xu, Y., Li, C., Zhang, K., and Batmanghelich, K.
\newblock Twin auxiliary classifiers gan.
\newblock In \emph{Advances in Neural Information Processing Systems}, volume~32, 2019.

\bibitem[Gong et~al.(2022)Gong, Li, Feng, Wu, and Kong]{gong2022diffuseq}
Gong, S., Li, M., Feng, J., Wu, Z., and Kong, L.
\newblock Diffuseq: Sequence to sequence text generation with diffusion models.
\newblock \emph{arXiv preprint arXiv:2210.08933}, 2022.

\bibitem[Goodfellow et~al.(2020)Goodfellow, Pouget-Abadie, Mirza, Xu, Warde-Farley, Ozair, Courville, and Bengio]{goodfellow2020generative}
Goodfellow, I., Pouget-Abadie, J., Mirza, M., Xu, B., Warde-Farley, D., Ozair, S., Courville, A., and Bengio, Y.
\newblock Generative adversarial networks.
\newblock \emph{Communications of the ACM}, 63\penalty0 (11):\penalty0 139--144, 2020.

\bibitem[Gu et~al.(2023)Gu, Gao, Zhai, Chen, Liu, and Susskind]{gu2023learning}
Gu, J., Gao, Q., Zhai, S., Chen, B., Liu, L., and Susskind, J.
\newblock Learning controllable 3d diffusion models from single-view images.
\newblock \emph{arXiv preprint arXiv:2304.06700}, 2023.

\bibitem[Gunasekar et~al.(2018{\natexlab{a}})Gunasekar, Lee, Soudry, and Srebro]{gunasekar2018characterizing}
Gunasekar, S., Lee, J., Soudry, D., and Srebro, N.
\newblock Characterizing implicit bias in terms of optimization geometry.
\newblock In \emph{International Conference on Machine Learning}, pp.\  1832--1841. PMLR, 2018{\natexlab{a}}.

\bibitem[Gunasekar et~al.(2018{\natexlab{b}})Gunasekar, Lee, Soudry, and Srebro]{gunasekar2018implicit}
Gunasekar, S., Lee, J.~D., Soudry, D., and Srebro, N.
\newblock Implicit bias of gradient descent on linear convolutional networks.
\newblock \emph{Advances in neural information processing systems}, 31, 2018{\natexlab{b}}.

\bibitem[Ho \& Salimans(2022)Ho and Salimans]{ho2022classifier}
Ho, J. and Salimans, T.
\newblock Classifier-free diffusion guidance.
\newblock \emph{arXiv preprint arXiv:2207.12598}, 2022.

\bibitem[Ho et~al.(2020)Ho, Jain, and Abbeel]{ho2020denoising}
Ho, J., Jain, A., and Abbeel, P.
\newblock Denoising diffusion probabilistic models.
\newblock \emph{Advances in neural information processing systems}, 33:\penalty0 6840--6851, 2020.

\bibitem[Ho et~al.(2022)Ho, Chan, Saharia, Whang, Gao, Gritsenko, Kingma, Poole, Norouzi, Fleet, et~al.]{ho2022imagen}
Ho, J., Chan, W., Saharia, C., Whang, J., Gao, R., Gritsenko, A., Kingma, D.~P., Poole, B., Norouzi, M., Fleet, D.~J., et~al.
\newblock Imagen video: High definition video generation with diffusion models.
\newblock \emph{arXiv preprint arXiv:2210.02303}, 2022.

\bibitem[Hochreiter \& Schmidhuber(1997)Hochreiter and Schmidhuber]{hochreiter1997long}
Hochreiter, S. and Schmidhuber, J.
\newblock Long short-term memory.
\newblock \emph{Neural computation}, 9\penalty0 (8):\penalty0 1735--1780, 1997.

\bibitem[Jain et~al.(2023)Jain, Choromanski, Singh, Sindhwani, Zhang, Tan, and Dubey]{jain2023mnemosyne}
Jain, D., Choromanski, K.~M., Singh, S., Sindhwani, V., Zhang, T., Tan, J., and Dubey, A.
\newblock Mnemosyne: Learning to train transformers with transformers.
\newblock \emph{arXiv preprint arXiv:2302.01128}, 2023.

\bibitem[Jiang et~al.(2018)Jiang, Chen, Shi, Dai, and Zhao]{jiang2018learning}
Jiang, H., Chen, Z., Shi, Y., Dai, B., and Zhao, T.
\newblock Learning to defend by learning to attack.
\newblock In \emph{arXiv preprint arXiv:1811.01213}, 2018.

\bibitem[Karras et~al.(2022)Karras, Aittala, Aila, and Laine]{karras2022elucidating}
Karras, T., Aittala, M., Aila, T., and Laine, S.
\newblock Elucidating the design space of diffusion-based generative models.
\newblock \emph{Advances in neural information processing systems}, 35:\penalty0 26565--26577, 2022.

\bibitem[Kingma \& Ba(2014)Kingma and Ba]{kingma2014adam}
Kingma, D.~P. and Ba, J.
\newblock Adam: A method for stochastic optimization.
\newblock \emph{arXiv preprint arXiv:1412.6980}, 2014.

\bibitem[Krishnamoorthy et~al.(2023)Krishnamoorthy, Mashkaria, and Grover]{krishnamoorthy2023diffusion}
Krishnamoorthy, S., Mashkaria, S.~M., and Grover, A.
\newblock Diffusion models for black-box optimization.
\newblock \emph{arXiv preprint arXiv:2306.07180}, 2023.

\bibitem[Langford \& Seeger(2001)Langford and Seeger]{langford2001bounds}
Langford, J. and Seeger, M.
\newblock \emph{Bounds for averaging classifiers}.
\newblock School of Computer Science, Carnegie Mellon University, 2001.

\bibitem[Li et~al.(2020)Li, Chen, You, Wang, and Lin]{li2020halo}
Li, C., Chen, T., You, H., Wang, Z., and Lin, Y.
\newblock Halo: Hardware-aware learning to optimize.
\newblock In \emph{European Conference on Computer Vision (ECCV)}, pp.\  500--518. Springer, 2020.

\bibitem[Li \& Malik(2016)Li and Malik]{li2016learning}
Li, K. and Malik, J.
\newblock Learning to optimize.
\newblock \emph{arXiv preprint arXiv:1606.01885}, 2016.

\bibitem[Liu et~al.(2022)Liu, Li, Du, Torralba, and Tenenbaum]{liu2022compositional}
Liu, N., Li, S., Du, Y., Torralba, A., and Tenenbaum, J.~B.
\newblock Compositional visual generation with composable diffusion models.
\newblock \emph{arXiv preprint arXiv:2206.01714}, 2022.

\bibitem[Lv et~al.(2017)Lv, Jiang, and Li]{lv2017learning}
Lv, K., Jiang, S., and Li, J.
\newblock Learning gradient descent: Better generalization and longer horizons.
\newblock In \emph{International Conference on Machine Learning (ICML)}, pp.\  2247--2255, 2017.

\bibitem[McAllester(1998)]{mcallester1998some}
McAllester, D.~A.
\newblock Some pac-bayesian theorems.
\newblock In \emph{Proceedings of the eleventh annual conference on Computational learning theory}, pp.\  230--234, 1998.

\bibitem[Metz et~al.(2022)Metz, Harrison, Freeman, Merchant, Beyer, Bradbury, Agarwal, Poole, Mordatch, Roberts, and Sohl-Dickstein]{metz2022velo}
Metz, L., Harrison, J., Freeman, C.~D., Merchant, A., Beyer, L., Bradbury, J., Agarwal, N., Poole, B., Mordatch, I., Roberts, A., and Sohl-Dickstein, J.
\newblock Velo: Training versatile learned optimizers by scaling up.
\newblock \emph{arXiv preprint arXiv:2211.09760v1}, 2022.

\bibitem[Odena et~al.(2017)Odena, Olah, and Shlens]{odena2017conditional}
Odena, A., Olah, C., and Shlens, J.
\newblock Conditional image synthesis with auxiliary classifier gans.
\newblock In \emph{International Conference on Machine Learning}, pp.\  2642--2651. PMLR, 2017.

\bibitem[Rombach et~al.(2022)Rombach, Blattmann, Lorenz, Esser, and Ommer]{rombach2022high}
Rombach, R., Blattmann, A., Lorenz, D., Esser, P., and Ommer, B.
\newblock High-resolution image synthesis with latent diffusion models.
\newblock In \emph{Proceedings of the IEEE/CVF conference on computer vision and pattern recognition}, pp.\  10684--10695, 2022.

\bibitem[Rumelhart et~al.(1986)Rumelhart, Hinton, and Williams]{rumelhart1986learning}
Rumelhart, D.~E., Hinton, G.~E., and Williams, R.~J.
\newblock Learning representations by back-propagating errors.
\newblock \emph{Nature}, 323\penalty0 (6088):\penalty0 533--536, 1986.

\bibitem[Schuetz et~al.(2022)Schuetz, Brubaker, and Katzgraber]{schuetz2022combinatorial}
Schuetz, M. J.~A., Brubaker, J.~K., and Katzgraber, H.~G.
\newblock Combinatorial optimization with physics-inspired graph neural networks.
\newblock \emph{arXiv preprint}, arXiv:2107.01188, 2022.

\bibitem[Shen et~al.(2021)Shen, Chen, Heaton, Chen, Liu, Yin, and Wang]{shen2021learning}
Shen, J., Chen, X., Heaton, H., Chen, T., Liu, J., Yin, W., and Wang, Z.
\newblock Learning a minimax optimizer: A pilot study.
\newblock In \emph{International Conference on Learning Representations (ICLR)}, 2021.

\bibitem[Song et~al.(2020)Song, Meng, and Ermon]{song2020denoising}
Song, J., Meng, C., and Ermon, S.
\newblock Denoising diffusion implicit models.
\newblock \emph{arXiv preprint arXiv:2010.02502}, 2020.

\bibitem[Song et~al.(2021)Song, Sohl-Dickstein, Kingma, Kumar, Ermon, and Poole]{song2021score}
Song, Y., Sohl-Dickstein, J., Kingma, D.~P., Kumar, A., Ermon, S., and Poole, B.
\newblock Score-based generative modeling through stochastic differential equations.
\newblock In \emph{International Conference on Learning Representations}, 2021.

\bibitem[Soudry et~al.(2018)Soudry, Hoffer, Nacson, Gunasekar, and Srebro]{soudry2018implicit}
Soudry, D., Hoffer, E., Nacson, M.~S., Gunasekar, S., and Srebro, N.
\newblock The implicit bias of gradient descent on separable data.
\newblock \emph{Journal of Machine Learning Research}, 19\penalty0 (70):\penalty0 1--57, 2018.

\bibitem[Vaswani et~al.(2017)Vaswani, Shazeer, Parmar, Uszkoreit, Jones, Gomez, Kaiser, and Polosukhin]{vaswani2017attention}
Vaswani, A., Shazeer, N., Parmar, N., Uszkoreit, J., Jones, L., Gomez, A.~N., Kaiser, {\L}., and Polosukhin, I.
\newblock Attention is all you need.
\newblock \emph{Advances in neural information processing systems}, 30, 2017.

\bibitem[Vicol et~al.(2021)Vicol, Metz, and Sohl-Dickstein]{vicol2021unbiased}
Vicol, P., Metz, L., and Sohl-Dickstein, J.
\newblock Unbiased gradient estimation in unrolled computation graphs with persistent evolution strategies.
\newblock In \emph{International Conference on Machine Learning}, pp.\  10553--10563. PMLR, 2021.

\bibitem[Wang et~al.(2024)Wang, Tang, Zeng, Yin, Xu, Zhou, Zang, Darrell, Liu, and You]{wang2024neural}
Wang, K., Tang, D., Zeng, B., Yin, Y., Xu, Z., Zhou, Y., Zang, Z., Darrell, T., Liu, Z., and You, Y.
\newblock Neural network diffusion.
\newblock \emph{arXiv preprint arXiv:2402.13144}, 2024.

\bibitem[Wang et~al.(2025)Wang, Tang, Zhao, and You]{wang2025recurrent}
Wang, K., Tang, D., Zhao, W., and You, Y.
\newblock Recurrent diffusion for large-scale parameter generation.
\newblock \emph{arXiv preprint arXiv:2501.11587}, 2025.

\bibitem[Xie et~al.(2024)Xie, Yin, and Wen]{xie2024ode}
Xie, Z., Yin, W., and Wen, Z.
\newblock Ode-based learning to optimize.
\newblock \emph{arXiv preprint arXiv:2406.02006v1}, 2024.

\bibitem[Xu et~al.(2023)Xu, Wang, Zhang, Wang, and Shi]{xu2023versatile}
Xu, X., Wang, Z., Zhang, G., Wang, K., and Shi, H.
\newblock Versatile diffusion: Text, images and variations all in one diffusion model.
\newblock In \emph{Proceedings of the IEEE/CVF International Conference on Computer Vision}, pp.\  7754--7765, 2023.

\bibitem[Zheng et~al.(2022)Zheng, Chen, Hu, and Wang]{zheng2022symbolic}
Zheng, W., Chen, T., Hu, T.-K., and Wang, Z.
\newblock Symbolic learning to optimize: Towards interpretability and scalability.
\newblock \emph{arXiv preprint arXiv:2203.06578}, 2022.

\end{thebibliography}
\bibliographystyle{icml2025}

\newpage
\appendix
\onecolumn

\section{Glossary}
\begin{table}[h]
\tablestyle{10pt}{1.3}
    \centering
\begin{tabular}{lll}
name
&
notation
&
comment

\\
\shline
solution
&
$x$
&
ground truth solutions
\\

trajectory
&
$\{x_{t}\}_{t\in[T_{\text{train}}]}$
&
ground truth trajectories, trained by optimizers

\\
blurred solution
&
$\tilde{x}_{t}$
&
solutions blurred by Gaussian noise

\\
blurred trajectory
&
$\{\tilde{x}_{t}\}_{t\in[T_{\text{blur}]}}$
&
trajectory blurred by Gaussian noise

\\
predicted solution
&
$\hat{x}_{t}$
&
generated by the backward diffusion process

\\
predicted trajectory
&
$\{\hat{x}_{t}\}_{t\in T_{\text{pred}}}$
&
predicted trajectory of diffusion process

\\
$\alpha$, $\beta$, $\gamma$
&
coefficients: SDE
&
time dependent, especially $\beta$ and $\gamma$

\\
$\mathbf{d}$
&
differentiate operator
&
conventional operator

\\
$\nabla$, $\nabla^{2}$
&
gradient and Hessian matrix operators
&
conventional operators

\\
$\dot{a}$, $\ddot{a}$
&
first and second order derivation of any a
&
conventional operators

\\
$u$, $v$
&
coefficients: time and Brownian motion
&
determining Wiener process (first order)

\\
$s, \sigma$
&
parameters: adjustment and intensity
&
determining general diffusion process

\end{tabular}
    \caption{Notations related in this paper.
    }
    \label{tab:appdx_glossary}
\end{table}

\begin{figure*}
    \centering
    \includegraphics[width=0.86\linewidth]{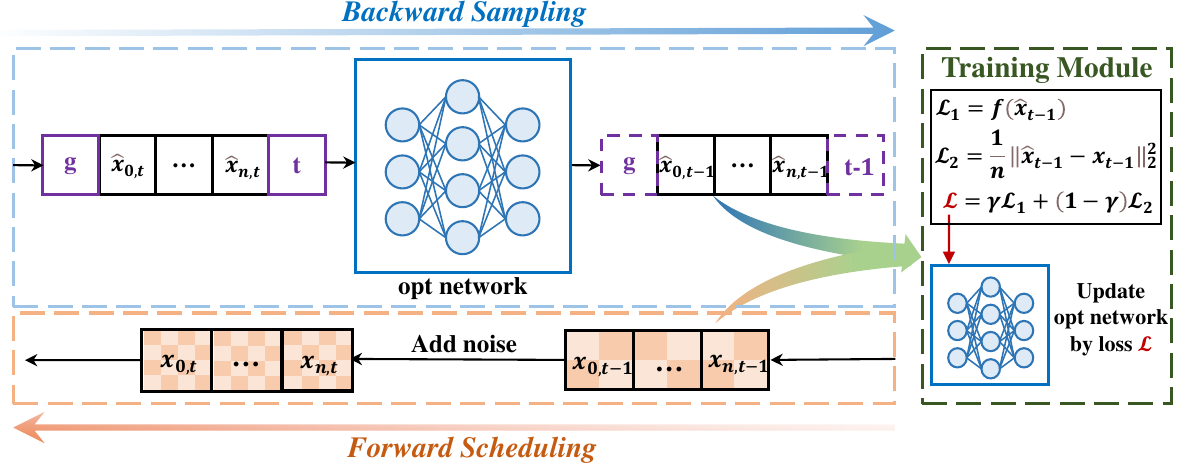}
    \caption{\small Model Training Framework for $\mathtt{Diff-L2O}$. The lower part is the trajectory generated by forward scheduling before training, and the upper part is the backward sampling from time step $t$ to $t-1$. Specifically, $\boldsymbol{{\hat{x}}_t}$ concatenated with guidance vector and time step embedding vector, is passed to the $\texttt{opt}$ network for one-step denoising. Based on $\boldsymbol{{\hat{x}}_{t-1}}$ and $\boldsymbol{{x}_{t-1}}$, we calculate the function loss for updating the $\texttt{opt}$ network. (The $x$ in this figure is the $\tilde{x}$ in the main paper.)}
    \vspace{-4mm}
    \label{fig:framework}
\end{figure*}

\section{Detailed Settings}
\subsection{Deep Neural Network on MNIST}
\paragraph{Model architectures.}
We consider the optimizee of MLPs with single hidden layer of dimension 20 and sigmoid activation function, using the cross-entropy loss on the MNIST dataset. 

\paragraph{Optimizees.}
\paragraph{Optimizees.} To evaluate our model, we deploy the following families of problems as the optimizees.

$\rhd$ \textit{Lasso.} We target to minimize the original LASSO objective function without considering the sparsity of the solution:
\begin{equation}
    \boldsymbol{x}^{Lasso} = \mathop{\arg\min}_{\boldsymbol{x}} \ \frac{1}{2}\|\mathbf{A}\boldsymbol{x}-\mathbf{b} \|_2^2 + \lambda \|\boldsymbol{x}\|_1
\end{equation}
where $\mathbf{A} \in \mathbb{R}^{n\times m}$ represent the characteristic matrix of a lasso problem instance, which is fixed and sampled from an \textit{i.i.d.} standard Gaussian distribution. $\mathbf{b} \in \mathbb{R}^{n \times 1}$ refers to the vector of dependent variables, which is also fixed and sampled from an \textit{i.i.d.} standard Gaussian distribution. $\lambda$ is the regularized hyperparameter set to $0.005$ in our experiment.

$\rhd$ \textit{Rastrigin.} Rastrigin is a common benchmark of non-convex optimization defined in $n$-dimensional space, where $n$ is the number of variables. It is characterized by a complex landscape of multiple local minima and a global minimum. We consider a family of Rastrigin function, and adopt the following definition from a seminal benchmark paper~\cite{chen2017learning}:
\begin{equation}
    \boldsymbol{x}^{Ras} = \mathop{\arg\min}_{\boldsymbol{x}} \ \frac{1}{2}\|\mathbf{A}\boldsymbol{x}-\mathbf{b} \|_2^2 - \alpha \mathbf{c}^\mathtt{T} \cos(2\pi\boldsymbol{x}) + \alpha n
\end{equation}
where $\mathbf{A} \in \mathbb{R}^{n \times n}$, $\mathbf{b} \in \mathbb{R}^{n \times 1}$ and $\mathbf{c} \in \mathbb{R}^{n \times 1}$ are all sampled from an \textit{i.i.d.} standard Gaussian distribution.

$\rhd$ \textit{Ackley.} Similar to Rastrigin function, Ackley function has many local minima which are comparably larger then the unique global minimum. Compare to Rastrigin, analytical optimizers can find the global minimum with less effort by enlarge their step-size. The problem is definded as:
\begin{equation}
    \boldsymbol{x}^{Ack} = \mathop{\arg\min}_{\boldsymbol{x}} \ 20 + e -20e^{-0.2 \|\mathbf{A}\boldsymbol{x}+\mathbf{b} \|_2} - e^{\frac{1}{n} \mathbf{c}^\mathtt{T} cos(2\pi\boldsymbol{x})}
\end{equation}
where $\mathbf{A} \in \mathbb{R}^{n \times n}$, $\mathbf{b} \in \mathbb{R}^{n \times 1}$ and $\mathbf{c} \in \mathbb{R}^{n \times 1}$ are all sampled from \textit{i.i.d.} standard Gaussian distributions.

\begin{figure}[h]
    \centering
    \includegraphics[width=0.45\linewidth]{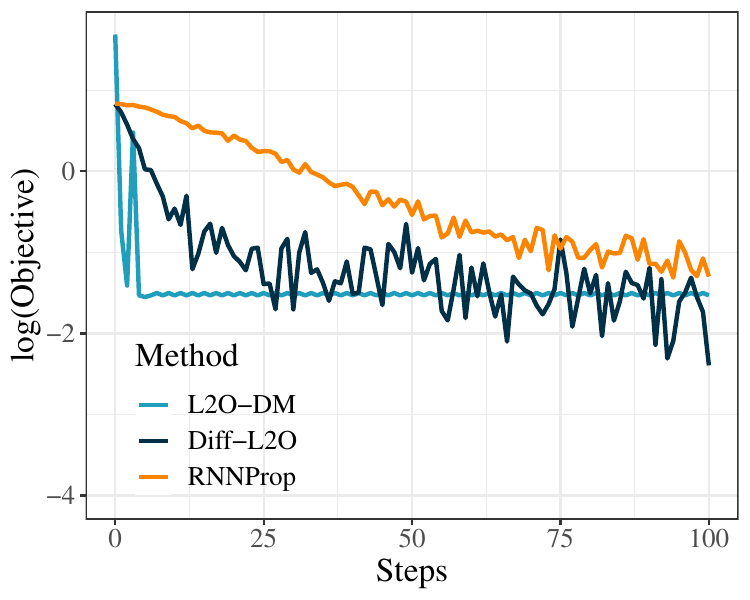}
    \caption{Comparison on MNIST.}
    \label{fig:comparison_mnist}
\end{figure}

\paragraph{Comparison: Loss Curves.} The loss curves between baselines and Diff-L2O are shown in Fig.~\ref{fig:comparison_mnist}.

\section{The element-wise variant of L2O}
Algorithm \ref{alg:mtd_diff_diffl2o_training} illustrate the Global-to-Local training philosophy by considering three phases, representing early, middle and later phase respectively. For each epoch, we first loop through each time step, and then loop through the positions, i.e. each element of the optimization variable. In early phase, we accumulate the training loss until the last element, called "Global"; In middle phase, we accumulate the training loss and conduct backward propagation on iterating every $\frac{d}{3}$ of elements, which is named Local. In the later phase, where we no longer accumulate the training loss, and this is when element-wise training is achieved.

\begin{algorithm}
\caption{Diff-L2O-ELE Training}
\label{alg:train3}
\textbf{Inputs:} $\hat{\boldsymbol{x}}_\mathtt{T}\sim\mathcal{N}(0, \mathbf{I})$, a guidance vector $\boldsymbol{g}$, its corresponding trajectory $\{\boldsymbol{x}_0,\boldsymbol{x}_1,\dots,\boldsymbol{x}_\mathtt{T}\}$, phase indicator $\mathtt{N}_1, \mathtt{N}_2$, dimension $d$
\begin{algorithmic}
\FOR{$n = 1, 2, \dots, \mathtt{N}$}
    \FOR{$t = \mathtt{T}, \mathtt{T}-1, \dots, 1$}
        \STATE $\boldsymbol{t} \gets \texttt{TE}(t)$
        \FOR{$\texttt{pos} = 1, 2, \dots, d$}
        \STATE $\boldsymbol{pos} \gets \texttt{PE}(\texttt{pos})$
        \STATE $\boldsymbol{x}_{t-1,pos} \gets  \texttt{opt}(\texttt{concat}(\boldsymbol{x}_t, \boldsymbol{g},\boldsymbol{t},\boldsymbol{pos}))$    
        \STATE $\mathcal{L}_1 \gets  f(\boldsymbol{\theta}, \hat{\boldsymbol{x}}_{t-1})$
        \STATE $\mathcal{L}_2 \gets \text{MSE}(\boldsymbol{x}_{t-1}, \hat{\boldsymbol{x}}_{t-1})$
        \STATE $\mathcal{L} \gets L + \gamma \mathcal{L}_1 + (1-\gamma) \mathcal{L}_2$
        \IF{$\mathtt{N} < \mathtt{N}_1$}
            \IF{$\texttt{pos} == d$}
            \STATE Update \texttt{opt} by minimizing $\mathcal{L}$ 
            \STATE $\mathcal{L} \gets 0$
            \ENDIF
        \ELSIF{$\mathtt{N}_1 \leq \mathtt{N} \le \mathtt{N}_2$}
            \IF{$\texttt{pos} \in { \lfloor \frac{d}{3} \rfloor, \lfloor \frac{2d}{3} \rfloor, \lfloor d \rfloor} $}
            \STATE Update \texttt{opt} by minimizing $\mathcal{L}$ 
            \STATE $\mathcal{L} \gets 0$
            \ENDIF
        \ELSE
            \STATE Update \texttt{opt} by minimizing $\mathcal{L}$ 
            \STATE $\mathcal{L} \gets 0$
        \ENDIF 
        \ENDFOR
    \ENDFOR
\ENDFOR
\end{algorithmic}
\end{algorithm}

\end{document}